\documentclass[sigconf]{acmart}
\acmSubmissionID{454}

\usepackage{booktabs}

\usepackage[ruled]{algorithm2e}

\SetAlFnt{\small}
\SetAlCapFnt{\small}
\SetAlCapNameFnt{\small}
\SetAlCapHSkip{0pt}

\copyrightyear{2023}
\acmYear{2023}
\setcopyright{rightsretained}
\acmConference[SIGGRAPH '23 Conference Proceedings]{Special Interest Group on Computer Graphics and Interactive Techniques Conference Conference Proceedings}{August 6--10, 2023}{Los Angeles, CA, USA}
\acmBooktitle{Special Interest Group on Computer Graphics and InteractiveTechniques Conference Conference Proceedings (SIGGRAPH '23 Conference Proceedings), August 6--10, 2023, Los Angeles, CA, USA}
\acmDOI{10.1145/3588432.3591517}
\acmISBN{979-8-4007-0159-7/23/08}

\begin{document}

\title{StyleAvatar: Real-time Photo-realistic Portrait Avatar from a Single Video}

\author{Lizhen Wang}
\orcid{0000-0002-6674-9327}
\affiliation{%
  \institution{Tsinghua University \& NNKosmos Technology}
  \city{Beijing \& Hangzhou}
  \country{China}
}
\email{wlz18@mails.tsinghua.edu.cn}

\author{Xiaochen Zhao}
\orcid{0000-0001-8976-7723}
\affiliation{%
  \institution{Tsinghua University}
  \city{Beijing}
  \country{China}
}
\email{zhaoxc19@mails.tsinghua.edu.cn}

\author{Jingxiang Sun}
\orcid{0000-0003-2966-9501}
\affiliation{%
  \institution{Tsinghua University}
  \city{Beijing}
  \country{China}
}
\email{sunjingxiang_stark@126.com}

\author{Yuxiang Zhang}
\orcid{0000-0002-8807-0825}
\affiliation{%
  \institution{Tsinghua University}
  \city{Beijing}
  \country{China}
}
\email{yx-z19@mails.tsinghua.edu.cn}

\author{Hongwen Zhang}
\orcid{0000-0001-8633-4551}
\affiliation{%
  \institution{Tsinghua University}
  \city{Beijing}
  \country{China}
}
\email{zhanghongwen@tsinghua.edu.cn}

\author{Tao Yu$^*$}
\orcid{0000-0002-3818-5069}
\affiliation{%
  \institution{Tsinghua University}
  \city{Beijing}
  \country{China}
}
\email{ytrock@mail.tsinghua.edu.cn}

\author{Yebin Liu$^*$}
\orcid{0000-0003-3215-0225}
\affiliation{%
  \institution{Tsinghua University}
  \city{Beijing}
  \country{China}
}
\email{liuyebin@mail.tsinghua.edu.cn}

\renewcommand\shortauthors{Wang, L. et al.}

\begin{abstract}
Face reenactment methods attempt to restore and re-animate portrait videos as realistically as possible. Existing methods face a dilemma in quality versus controllability: 2D GAN-based methods achieve higher image quality but suffer in fine-grained control of facial attributes compared with 3D counterparts. In this work, we propose StyleAvatar, a real-time photo-realistic portrait avatar reconstruction method using StyleGAN-based networks, which can generate high-fidelity portrait avatars with faithful expression control. 
We expand the capabilities of StyleGAN by introducing a compositional representation and a sliding window augmentation method, which enable faster convergence and improve translation generalization. Specifically, we divide the portrait scenes into three parts for adaptive adjustments: facial region, non-facial foreground region, and the background. Besides, our network leverages the best of UNet, StyleGAN and time coding for video learning, which enables high-quality video generation. Furthermore, a sliding window augmentation method together with a pre-training strategy are proposed to improve translation generalization and training performance, respectively.
The proposed network can converge within two hours while ensuring high image quality and a forward rendering time of only 20 milliseconds. Furthermore, we propose a real-time live system, which further pushes research into applications. Results and experiments demonstrate the superiority of our method in terms of image quality, full portrait video generation, and real-time re-animation compared to existing facial reenactment methods. Training and inference code for this paper are at https://github.com/LizhenWangT/StyleAvatar.
\end{abstract}




\begin{teaserfigure}
\centering
  \includegraphics[width=\textwidth]{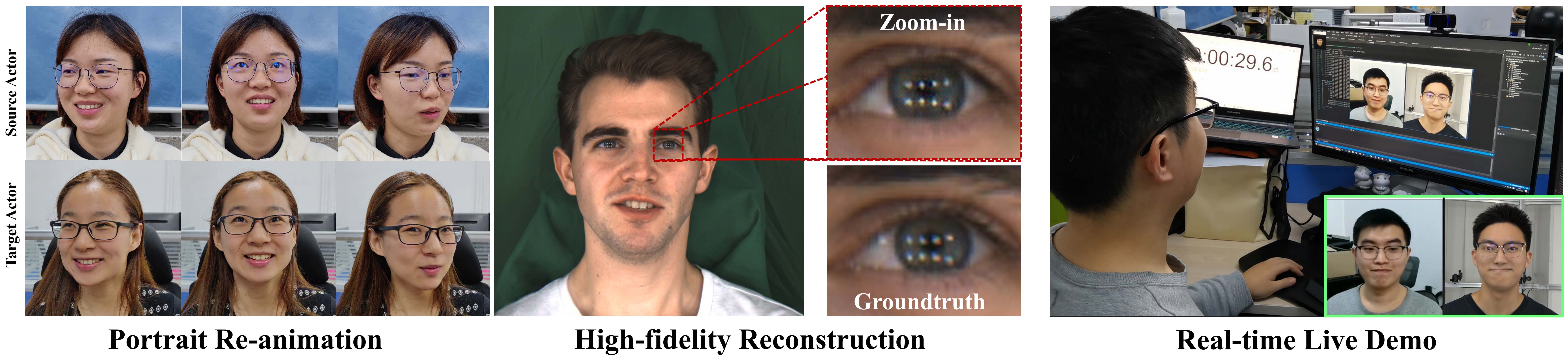}
  \caption{We present StyleAvatar, a real-time high-fidelity portrait avatar reconstruction method.}
  \Description{We present StyleAvatar, a real-time high-fidelity portrait avatar reconstruction method.}
  \label{fig:teaser}
\end{teaserfigure}

\maketitle

\def\thefootnote{*}\footnotetext{Corresponding authors}\def\thefootnote{\arabic{footnote}}

\section{Introduction}
\label{sec:intro}

Photo-realistic portrait avatar reconstruction and re-animation is a long standing topic in computer vision and computer graphics, with a wide range of applications from video editing to mixed reality. Recent efforts on portrait avatar based on NeRF~\cite{nerface, Gao2022nerfblendshape, zheng2023avatarrex} or some other 3D representations present~\cite{zheng2022pointavatar, zheng2022avatar, zielonka2022instant, grassal2021neuralhead} have demonstrated that stable 3D avatars can be learned from monocular videos. However, compared to 2D GAN-based methods, most 3D approaches still face limitations in resolution and image quality. Additionally, these methods primarily focus on the rigid facial regions, ignoring long hair, body parts, and background elements. While backgrounds can be overlaid directly, poorly executed overlays often lead to unrealistic results. The ideal portrait avatar should prioritize high-fidelity, fast training, fine-grained control, and real-time efficiency.

Learning 3D head avatars from monocular videos has been a popular topic in recent years. Early works ~\cite{nerface, zheng2022avatar, grassal2021neuralhead} incorporate NeRF into head avatars, achieving promising view consistency. More recent approaches either aim to achieve even better rendering realism~\cite{zheng2022pointavatar, xu2023latentavatar} or faster training convergence and inference speed~\cite{xu2023avatarmav, zielonka2022instant, Gao2022nerfblendshape} by utilizing more efficient 3D representations~\cite{mueller2022instantngp, TiNeuVox}. In general, the core idea of 3D methods is to maintain a relatively fixed feature space, such as topology-consistent meshes or a canonical space, to enable each point or voxel to learn certain local features from the video. This strategy leads to greater stability, but also results in smoothed textures due to tracking instability or other factors.

On the other end of the spectrum, benefiting from the powerful StyleGAN~\cite{karras2020a, karras2020analyzing, styleganv3}, following works~\cite{tewari2020stylerig, abdal2020styleflow, tewari2020pie, Wang_2021_CVPR, deng2020disentangled, shen2020interfacegan, harkonen2020ganspace, sofgan} have drastically improved the semantic editing performance. Some methods~\cite{doukas2021headgan, Drobyshev22MP, Khakhulin2022ROME} can create head avatars from a single image, while others~\cite{xiang2022gram, sun2022next3d, sun2022ide} can even produce controllable 3D faces with EG3D~\cite{Chan2022eg3d}. Nevertheless, these StyleGAN-based methods mostly rely on a highly aligned HD face dataset such as FFHQ, which lacks sufficient variation in facial expressions. Additionally, they can not generate natural head movements in portrait videos.

We propose StyleAvatar, a real-time system for photo-realistic portrait avatar reconstruction using a StyleGAN-based network. Our system is capable of generating a high-fidelity portrait avatar in just three hours. To address the challenges of full photo-realistic portrait video reconstruction, we divide the portrait scenes into three parts: the face, movable body parts (shoulders, neck, and hair), and background. Each part has distinct attributes: the face part provides almost all the moving information through 3DMM, while the background is typically static. The movable body part may contain numerous uncontrollable movements, but we can still learn some trends from facial movements.

To overcome these challenges, we propose StyleAvatar, a real-time system for photo-realistic portrait avatar reconstruction using a StyleGAN-based network. Our system can generate a high-fidelity portrait avatar in just two hours. In order to reconstruct a full portrait video, we divide the video into three parts: facial region, non-facial foreground region (shoulders, neck, hair etc.) and background. Each part has distinct attributes: the facial region can be described by the 3DMM; the non-facial foreground region often exhibits uncontrollable movements, but trends can be learned from facial movements; and the background remains static. To better represent the distinct features of the three parts, we use two StyleGAN generators to generate two static feature maps for the facial region and background, and propose a StyleUNet to generate the non-facial foreground feature map from the input 3DMM rendering. 
To accelerate the training and inference speed, we use use Neural Textures~\cite{Thies2019DeferredNR} for the facial region during the feature combination stage. Moreover, a sliding window augmentation method is introduced to improve translation generalization and we pre-train the model on a small video dataset to further speed up training. Finally, another StyleUNet is used to generate the final images from the combined feature maps. Our framework is designed to be easily accelerated by TensorRT and OpenGL, with a forward rendering time of only 20 milliseconds, enabling real-time live portrait reenactment. Results and experiments demonstrate that our method outperforms existing facial reenactment methods in terms of image quality, full portrait video generation, and real-time re-animation. Our contributions can be concluded as:
\begin{itemize}
\item We introduce a compositional representation that effectively decomposes the facial region, the non-facial foreground region and the background, so that we can make adaptive adjustments according to the characteristics of different regions to increase the stability and the training speed.
\item We further propose StyleUNet that leverages the best of UNet, StyleGAN and time coding for video learning, which enables high-quality video generation.
\item A sliding window augmentation method together with a pre-training strategy are proposed to improve translation generalization and training performance, respectively.
\end{itemize}

\begin{figure*}
	\begin{center}
		\includegraphics[width=\linewidth]{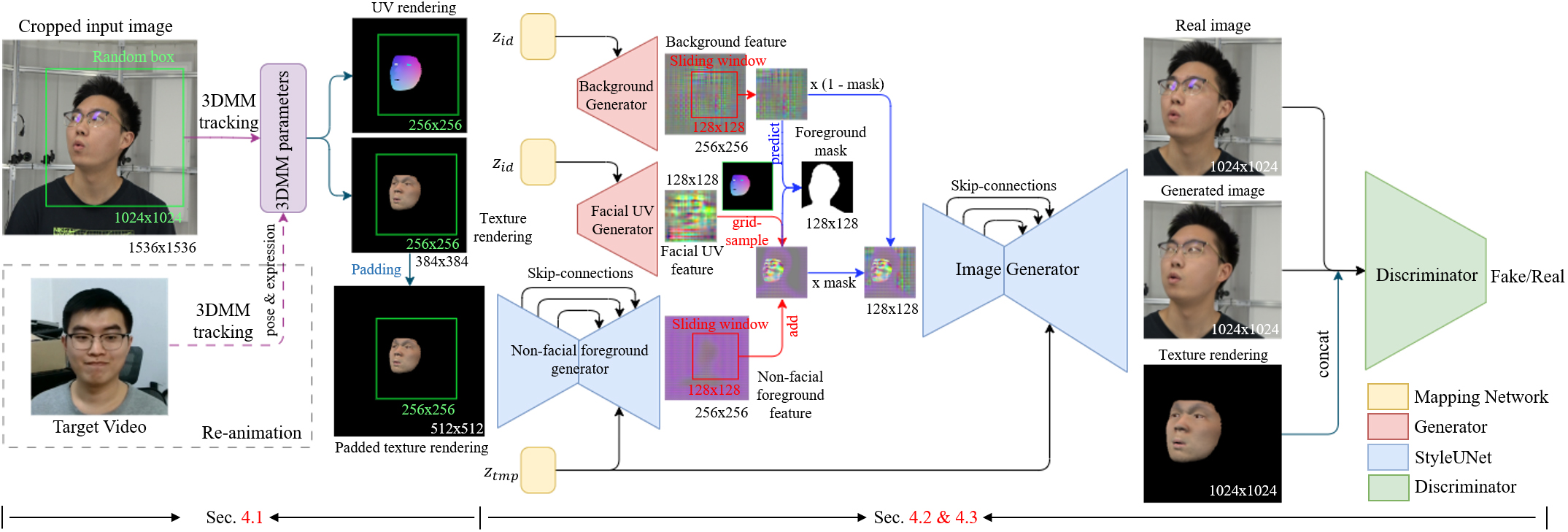}
	\end{center}
\caption{Our portrait avatar reconstruction and re-animation pipeline consists of three main steps: 3DMM tracking and rendering, feature generation of the facial region, non-facial foreground region, and background, and final image generation from the combined feature map. To achieve this, we utilize two StyleGAN generators, a StyleGAN discriminator, and two StyleUNets. Additionally, we incorporate data augmentation techniques with random boxes on input images and sliding windows on generated feature maps to improve translation generalization.}
\label{fig:pipeline}
\end{figure*}

\section{Related Works}
\label{sec:related}

%Our method was inspired by both video-based facial reenactment methods and StyleGAN based facial image generation and editing methods. So we divide the related work into two sections. 

\paragraph{Facial Reenactment Methods.} These methods aim at generating photo-realistic portrait images (including face, hair, neck and even shoulder regions) of a target person given the performance of another person, which is different from face replacement methods~\cite{deepfacelab} or face performance capture and animation methods~\cite{weise2011realtime, li2012data}. According to the input data, facial reenactment methods can be roughly divided into three categories: multi-view system based methods, single video based methods and single image based methods. Based on multi-view capture systems, recent researches~\cite{lombardi2018deep, wei2019vr, ma2021pixel, raj2021pva, lombardi2019neural, Wang2022MoRFMR} were able to generate facial avatars with impressive subtle details and highly flexible controllability for immersive metric-telepresence. However, difficulties in data acquisition limited the broad applications. On contrast, single image-based methods~\cite{liu2001expressive, vlasic2005face, olszewski2017realistic, averbuch-elor2017bringing, siarohin2019first, geng2019warp, nagano2019pagan, doukas2021headgan, kowalski2020config, mallya2022implicit, yin2022styleheat, Drobyshev22MP, hong2022depth} were most easy to capture and could produce photo-realistic facial reenactment results. However, shapes and details may not be consistent when animating to large poses and expressions especially for those regions that have not been well covered in the single input image, not to say the dynamic facial details. Single video based methods~\cite{garrido2014automatic, thies2015real, thies2016face2face, suwajanakorn2017synthesizing, doukas2021head2head, koujan2020head2head} showed more stable facial reenactment results. Among them, ~\cite{kim2018deep} presented impressive full head reenactment and interactive editing results based on an image2image translation framework. Recent methods~\cite{nerface, wang2021learning, Gao2022nerfblendshape, zielonka2022instant, xu2023avatarmav, xu2023latentavatar} showed state-of-the-art reenactment results with a parameter-controlled neural radiance field. Other methods~\cite{grassal2022neural, cao2022authentic, zheng2022avatar, zheng2022pointavatar, garbin2022voltemorph} improved stability and texture quality by utilizing 3D representations like meshes or point clouds. However, existing single-video-based facial reenactment methods still face challenges in recovering elaborate details such as hairs, freckles, and even skin pores.

\paragraph{StyleGAN-based Facial Image Generation and Editing Methods.} These methods can produce high-resolution and photo-realistic facial images ~\cite{karras2020a,karras2020analyzing,styleganv3} even under semantic editing operations~\cite{tewari2020stylerig, abdal2020styleflow, tewari2020pie, ghosh2020gif, shen2020interpreting, Wang_2021_CVPR, deng2020disentangled, shen2020interfacegan, harkonen2020ganspace, sofgan, shi2022semanticstylegan, richardson2021encoding, tov2021designing, alaluf2021restyle, ren2021pirenderer, jang2021stylecarigan}. These works conducted semantic modifications on the generated images of StyleGAN by decoupling the input latent space. Moreover, Ghosh \textit{et, al.}~\cite{ghosh2020gif} and Shoshan \textit{et, al.}~\cite{shoshan2021gan} combined the StyleGAN latent space with the semantic input, such as 3DMM parameters~\cite{CGIT1999Blanz}. SofGAN~\cite{sofgan} used semantic segmentation maps to condition the image generator and achieves 3D aware generation using 3D scans additionally for training. These methods provide meaningful control over the shape, pose, hair and style of the generated photo-realistic facial images, but fine-grained expression modifications such as blink remain unavailable. More importantly, these methods heavily rely on a pre-trained StyleGAN latent space, making it difficult for them to maintain temporal stability and consistency of details during the editing process.

\section{Method}
\label{sec:method}

As shown in Fig.~\ref{fig:pipeline}, input with a monocular portrait video, we first perform 3DMM tracking (Sec.~\ref{sec:tracking}) to generate synthetic renderings with both predicted texture and UV coordinate vertex colors. Next, in the feature generation stage (Sec.~\ref{sec:feature_gen}), we divide a portrait feature map into three parts: a static facial feature map generated in UV space generated by a StyleGAN generator; a non-facial foreground feature map generated from the input texture rendering by a StyleUNet; a static background feature map generated by another StyleGAN generator. Then, we employ Neural Textures to extract a facial feature map from the UV space, and introduce a sliding window data augmentation during the combination of feature maps to better utilize information from the entire video. Finally, another StyleUNet is used to generate images from the combined feature maps and a StyleGAN discriminator is introduced for adversarial learning. Note that only the two StyleUNets are computed in the inference stage, and re-animation can be achieved by replacing the input expression and pose parameters from another video.

\subsection{Data Processing}
\label{sec:tracking}

We first perfrom 3DMM tracking on the input monocular portrait video to generate pixel-aligned 3DMM renderings for subsequent training. We choose to use FaceVerse~\cite{wang2022faceverse} due to its rich shape and expression bases and we add separate eyeballs. A texture rendering with predicted texture is used for non-facial foreground feature generation and another UV rendering with UV coordinate vertex colors is used for the Neural Textures of the facial region. To supervise the training of mask prediction in the feature combination stage, we generate foreground masks using Robust Video Matting~\cite{rvm}.

For the 3DMM tracking algorithm, we need to solve for shape coefficients $\theta_{shape}$, expression coefficients $\theta_{expression}$, texture coefficients $\theta_{texture}$, translation $t$, scale $s$, and rotations of the head and two eyeballs $R_1, R_2, R_3$. To improve efficiency, we directly solve for the analytical solutions of these parameters from facial landmarks ${K}_{tgt}$ detected by MediaPipe\footnote{https://github.com/google/mediapipe}. Specifically, we utilize the following energy functions:
\begin{equation}
\small
\underset{R_x,t,s}{\arg \min}\|R_x{K}_{src} + t - {K}_{tgt}/s\|_2
 \label{eq:rotation} 
\end{equation}
\begin{equation}
\small
\underset{\delta\theta_{shape},\delta\theta_{exp}}{\arg \min}\|R_1({K}_{src} + B_{shape}\delta\theta_{shape} + B_{exp}\delta\theta_{exp}) + t - {K}_{tgt}/s\|_2
 \label{eq:coeff} 
\end{equation}
\begin{equation}
\small
\underset{\delta\theta_{tex}}{\arg \min}\|{T}_{src} + B_{tex}\delta\theta_{tex} - {T}_{tgt}\|_2
 \label{eq:tex} 
\end{equation}
where ${K}{src}$ represents the corresponding landmarks on FaceVerse, while $B{shape}, B_{exp}, B_{tex}$ represent the shape, expression, and texture bases. To obtain the final parameters, we solve Eq.~\ref{eq:rotation} and Eq.~\ref{eq:coeff} iteratively. Eq.~\ref{eq:rotation} provides us with the rotation, translation, and scale parameters, whereas Eq.~\ref{eq:coeff} gives us the shape and expression coefficients of FaceVerse. Additionally, we use Eq.~\ref{eq:tex} to solve for the texture coefficients. Eq.~\ref{eq:coeff} and Eq.~\ref{eq:coeff} can be solved by LDLT decomposition, while Eq.~\ref{eq:rotation} can be solved by SVD decomposition. Since we require the predicted shape to be consistent for the same person, and the predicted texture is not critical for our framework, we only solve for the shape and texture coefficients in the first frame.

To generate the UV rendering, we use UV coordinate vertex colors, where the red and green channel pixel values correspond to the x and y position in the UV coordinate. To generate the texture rendering, we utilize the predicted texture and render it with a pre-defined fixed lighting in order to accentuate facial expression changes. Furthermore, in order to better utilize information from the entire video, we propose a sliding window data augmentation approach. Specifically, we enlarge the crop box of the portrait region by 1.5 times, allowing for the preservation of more upper body and background information, as shown in Fig.~\ref{fig:pipeline}.

\subsection{Feature Generation and Combination}
\label{sec:feature_gen}

In the feature generation stage, to create high-quality avatars, we divide the portrait region into three parts and employ StyleGAN-based networks to generate corresponding feature maps. Specifically, we use two StyleGAN generators to generate static feature maps for the facial region and background. The facial feature maps are generated in UV space, and the input identity latent code $z_{id}$ is designed for pre-training, allowing us to generate different feature maps for different videos. For the non-facial foreground region, we propose a StyleUNet to generate the pixel-aligned feature map from the input 3DMM rendering. We believe that plausible movements of this part can be predicted from the movements of 3DMM. To accommodate uncontrollable changes like hair motion, we use the input temporal latent code $z_{tmp}$ as an additional input. We follow Bahmani et al.~\cite{bahmani20223d} in using positional embedding to map person identities and timestamps to higher dimensions, so that they can be input into our network.

\begin{figure}
	\begin{center}
		\includegraphics[width=0.9\linewidth]{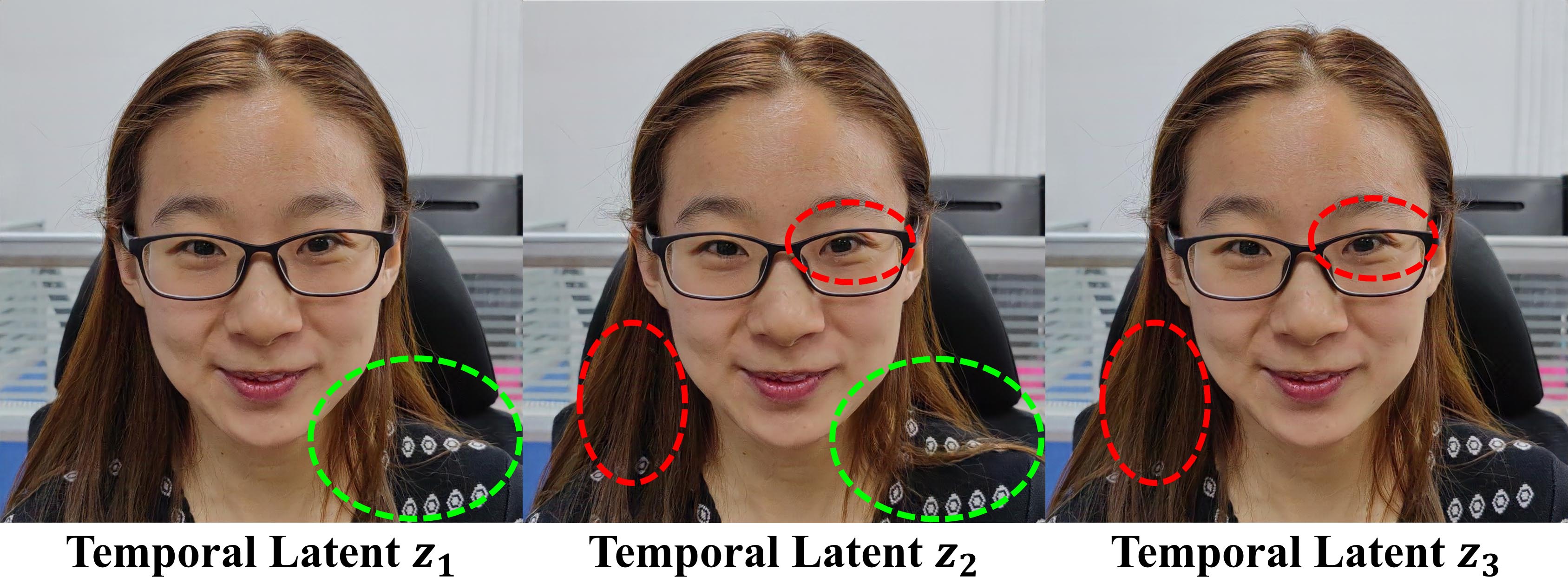}
	\end{center}
\caption{The information stored in our temporal latent space allows for changes in details such as hair changes when inputting the same rendering with different temporal latent code $z_{tmp}$.}
\label{fig:method2}
\end{figure}

In the feature combination stage, we combine the three feature maps to generate the final image. For the facial region, the 3DMM already provides the basic geometry, so we adopt the Neural Textures proposed in Deferred Neural Rendering (DNR)~\cite{Thies2019DeferredNR} to generate a facial feature map by sampling from the facial UV feature map using the UV rendering. As for the background and non-facial foreground regions, we crop the generated feature maps using sliding windows to ensure the pixel-aligned relationship between the input feature maps and the output images. To generate dynamic facial changes and protruding shapes such as glasses, we directly add the facial feature map and the non-facial foreground feature map. To ensure that the foreground feature map is in front of the background feature map, we first use a supervised convolutional layer to predict a foreground mask from the foreground and background feature maps, and then combine these two feature maps using the predicted mask. Finally, we use another StyleUNet to generate the final image from the combined feature map and the networks are trained adversarially with a StyleGAN discriminator. We also incorporate the temporal latent code $z_{tmp}$ as an additional input to our StyleUNet. As depicted in Fig.~\ref{fig:method2}, varying $z_{tmp}$ values result in time-related changes such as hair movement. The use of a mapping network and $z_{tmp}$ helps prevent overly smoothed details by incorporating time-related but uncontrollable changes into the latent space. Note that only the two StyleUNets will be computed during the inference stage.

\begin{figure}
	\begin{center}
		\includegraphics[width=\linewidth]{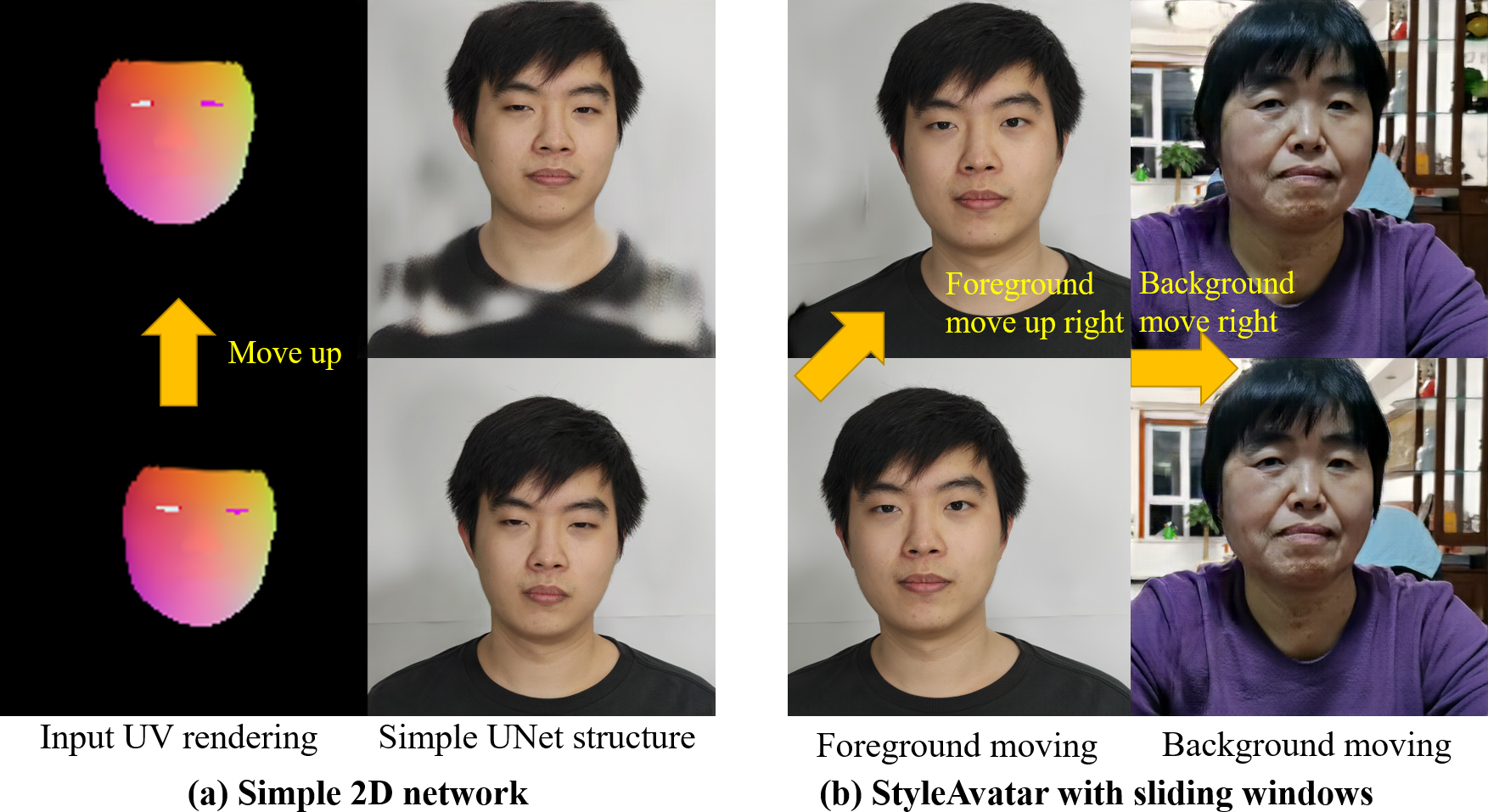}
	\end{center}
\caption{Free movement in a large range may tear the generated image using a simple UNet. By introducing the sliding windows our avatar can move in a larger range.}
\label{fig:method1}
\end{figure}

To overcome the challenge of handling free translational motions in 2D avatars, we propose a sliding window data augmentation method to better utilize the information from the entire input video. As demonstrated in Fig.~\ref{fig:method1}a, the use of a simple UNet-based 2D avatar often leads to noticeable visual artifacts, such as image tearing, particularly in cases of excessive head translations. This can be attributed to two factors: firstly, UNets are known to be highly sensitive to image translations; secondly, the central area of the input portrait video tends to have more pronounced pose and expression changes, whereas the border area shows less variation, leading to a reduced ability to generalize translations. 
To enhance the translation generalization ability, we first employ random sample boxes on both the input images and their corresponding 3DMM renderings. To ensure pixel-aligned features in the non-facial foreground region, we pad the sampled input rendering to a higher resolution (256 to 512). Then, to extract pixel-aligned background and non-facial foreground regions from the generated feature maps, we utilize two sliding windows. The position of these sliding windows is determined by the input random sample boxes. Finally, we concatenate the pixel-aligned texture renderings with the generated images to create fake input images for our discriminator, and concatenate the texture renderings with the input images inside the sample boxes to create real input images for our discriminator. As shown in Fig.~\ref{fig:method1}b, with the aid of the proposed data augmentation method, our approach is capable of addressing large head translations and can generate both foreground and background movements by employing different sliding windows during feature combination.

\subsection{Network Structure and Loss Functions}
\label{sec:network}

As shown in Fig.~\ref{fig:pipeline}, our framework consists of five networks: two feature generators, two StyleUNets, and a discriminator. To meet real-time requirements and speed up training, we employ wavelet transform in all networks similar to SWAGAN~\cite{gal2021swagan}, which enables us to replace a $1024\times1024\times3$ image with a $512\times512\times12$ representation. Apart from this, the remaining parts of our generator and discriminator follow StyleGAN2 architecture. The StyleUNet uses an encoder-decoder structure, designed for image-to-image translation, with additional input latent code and noise.
Similar to UNet~\cite{ronneberger2015u-net:}, our encoder extracts multi-scale features that are sent to the decoder via skip-connections. We use a mapping network, the modulated convolution and the temporal latent code to introduce additional variation possibilities for the network, preventing it from smoothing certain features. For example, the uncontrolled hair motions can be accommodated by the time-related latent code. However, this approach cannot guarantee the complete stability of the hair. We use the 64-dimensional latent code as the input of mapping networks in both StyleUNets and generators. As described in~\cite{styleganv3}, the noise injection layers of StyleGAN2 can lead to texture sticking artifacts. To mitigate this effect, we use UV renderings to map the noise from UV space to the face regions, and a fixed noise is used for the static background. To incorporate facial priors into the discriminator, we concatenate the texture rendering with the output image, which serves as the input to the discriminator.

In terms of loss functions, as direct supervision is feasible for our task, we utilize common L1 loss and perceptual loss with a VGG19 during the training process. Additionally, we incorporate an L1 loss for the foreground mask. We also include GAN loss for adversarial learning. Our loss functions can be formulated as
\begin{equation}
\label{equ:loss}
\begin{aligned}
\small
\mathcal{L} = \mathcal{L}_{1} + \mathcal{L}_{percep} + \mathcal{L}_{mask} + \mathcal{L}_{GAN}
\end{aligned}
\end{equation}
%\begin{equation}
%\label{equ:ad loss}
%\begin{aligned}
%\small
%\mathcal{L}_{GAN} = \mathbb{E}_{I_{gt}}[log \mathcal{D}(I_{gt}, R)] + \mathbb{E}_{I_{out}}[log (1-\mathcal{D}(I_{out}, R))] 
%\end{aligned}
%\end{equation}

\subsection{Pre-training and Live System}
\label{sec:live system}

To expedite training convergence, we use 6 videos cropped from 4K videos for pre-training. As discussed in Sec.~\ref{sec:feature_gen}, we use distinct identity latent codes $z_{id}$ for each video. Despite the limited dataset, pre-training has proven effective, as demonstrated in Sec.~\ref{sec:ablation}. Note that we assume a fixed video length for $z_{tmp}$ of each video and select a fixed $z_{tmp}$ during the inference stage. %The pre-trained model already has latent training logic pre-loaded, such as how to learn the foreground mask from the input features, which determines the direction of gradient propagation.

To bring our work closer to practical applications, we present a real-time live system consisting of three main steps: 3DMM tracking, OpenGL rendering for input 3DMM renderings, and the two StyleUNets which have been converted to TensorRT models. Our system can run at 35 fps (28 ms per frame) using a 16-bit TensorRT model on a PC with one RTX 3090 GPU, and requires approximately 4 GB of GPU memory during the inference stage. The 16-bit model takes an average of 20 milliseconds GPU time, while the original PyTorch model takes an average of 31 milliseconds. To generate a realistic facial avatar, we need approximately two minutes of video footage featuring a range of head poses and facial expressions. The videos used in this paper include the Obama video courtesy of the White House (public domain), a video provided by IMAvatar, and videos from the MEAD dataset~\cite{kaisiyuan2020mead}. We obtained consent from all actors featured in the remaining videos.

\section{Experiments}
\label{sec:experiments}

\subsection{Comparisons}
\label{sec:compare}

\begin{figure}
	\begin{center}
		\includegraphics[width=\linewidth]{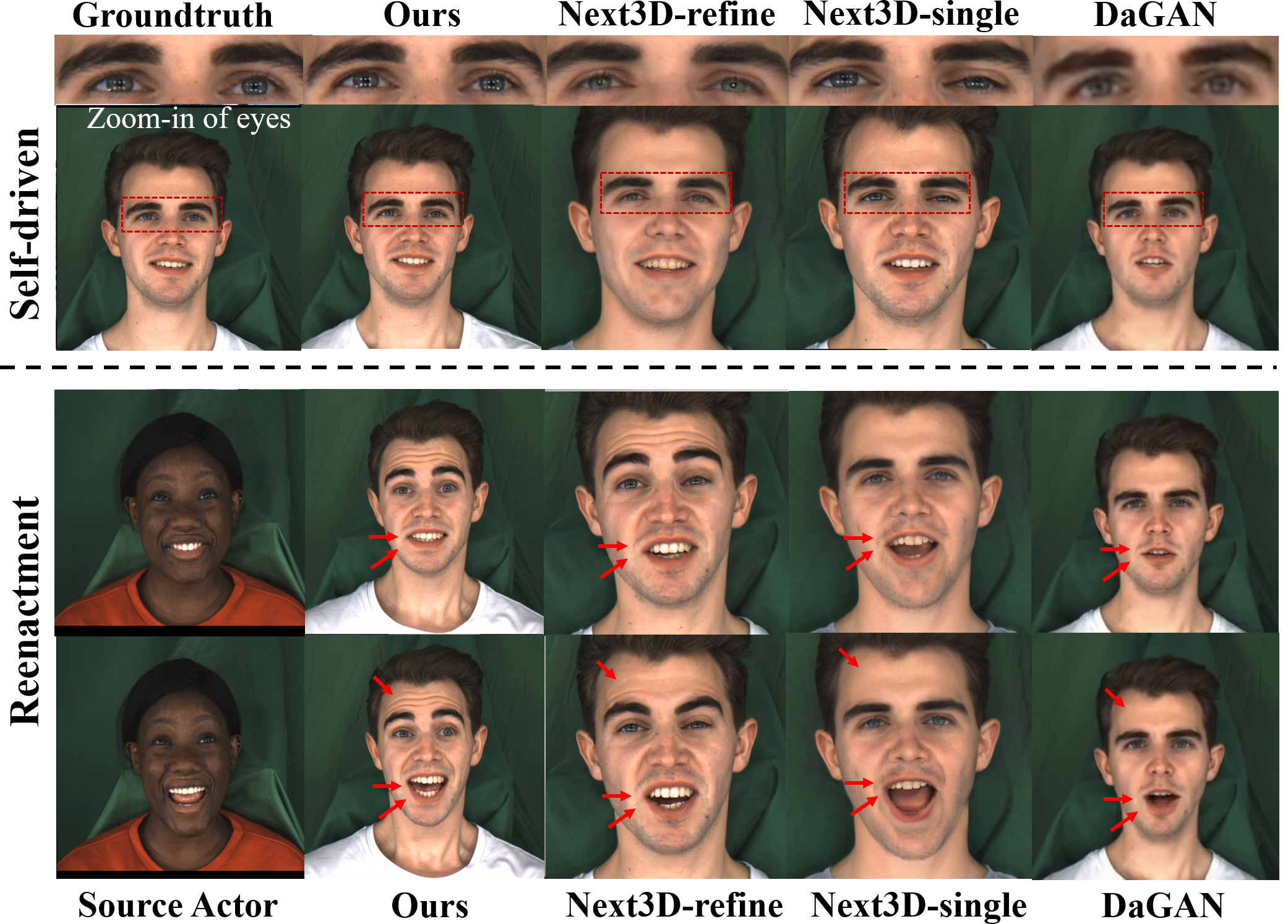}
	\end{center}
\caption{Self-driven animation and reenactment comparisons with one-shot portrait avatar methods.}
\label{fig:exp1}
\end{figure}

\begin{table}
\small
\begin{center}
\caption{Quantitative comparisons with one-shot avatar methods.}
    \begin{tabular}{lccc}
    \toprule
Error Metric            & SSIM $\uparrow$  & PSNR $\uparrow$  & FID $\downarrow$ \\ \midrule
DaGAN             & 0.79 & 22.0 & 73.2   \\
Next3D-single     & 0.81 & 24.6 & 24.4  \\
Next3D-refine      & 0.79 & 23.5 & 21.4   \\
Ours            & $\mathbf{0.87}$ & $\mathbf{27.1}$ & $\mathbf{12.2}$  \\ \bottomrule
    \end{tabular}
%\newline
\label{tab:tab1}
\end{center}
\end{table}

We first compare our method to state-of-the-art one-shot facial methods based on the GAN structure to demonstrate our network's ability to achieve fine-grained control of expressions and facial details. For comparison, we select two representative approaches: DaGAN~\cite{hong2022depth} as a representation of 2D one-shot avatars and Next3D~\cite{sun2022next3d} as a representation of 3D one-shot methods.
To ensure a fair experiment, we fine-tune Next3D on the monocular video used for the comparison for 24 hours, denoted as ``Next3D-refine'', while ``Next3D-single'' refers to the original pre-trained model. As shown in Fig.~\ref{fig:exp1} and Tab.~\ref{tab:tab1}, our method outperforms DaGAN and both versions of Next3D in both image quality and control of facial attributes. Although Next3D-refine improves fidelity after fine-tuning on the video, it still cannot achieve fine-grained control of facial expressions. These results suggest that video-based training is still necessary for high-fidelity portrait avatars, and our network structure shows better performance in video-based training. Note that in order to obtain the values presented in Tab.~\ref{tab:tab1}, we have first aligned the faces, cropped the images, removed the background, and resized them to a resolution of $512\times512$. The values presented in Tab.~\ref{tab:tab1} are calculated based on self-driving images generated from the testing set, which is another video in the MEAD dataset, and the corresponding ground-truth images.

\begin{table}
\small
\begin{center}
\caption{Quantitative comparisons with video-based avatar methods.}
    \begin{tabular}{lcccccc}
    \toprule
    & \multicolumn{3}{c}{Case 1} & \multicolumn{3}{c}{Case 2}\\
Error Metric    & SSIM $\uparrow$  & PSNR $\uparrow$  & FID $\downarrow$ & SSIM $\uparrow$  & PSNR $\uparrow$  & FID $\downarrow$ \\ \midrule
DVP      & 0.80 & 21.6 & 31.3 & 0.85 & 24.2 &  25.9  \\
NeRFace  & 0.77 & 18.8 & 36.4 & 0.87 & 22.2 & 50.0   \\
NHA      & 0.73 & 16.0 & 83.5 & 0.86 & 19.5 & 62.9   \\
IMAvatar & 0.78 & 19.0 & 59.7 & $\mathbf{0.89}$ & 25.6 & 58.7   \\
Ours     & $\mathbf{0.87}$ & $\mathbf{25.6}$ & $\mathbf{15.1}$ & 0.87 & $\mathbf{26.2}$ & $\mathbf{13.2}$ \\ \bottomrule
    \end{tabular}
%\newline
\label{tab:tab2}
\end{center}
\end{table}

\begin{figure}
	\begin{center}
		\includegraphics[width=\linewidth]{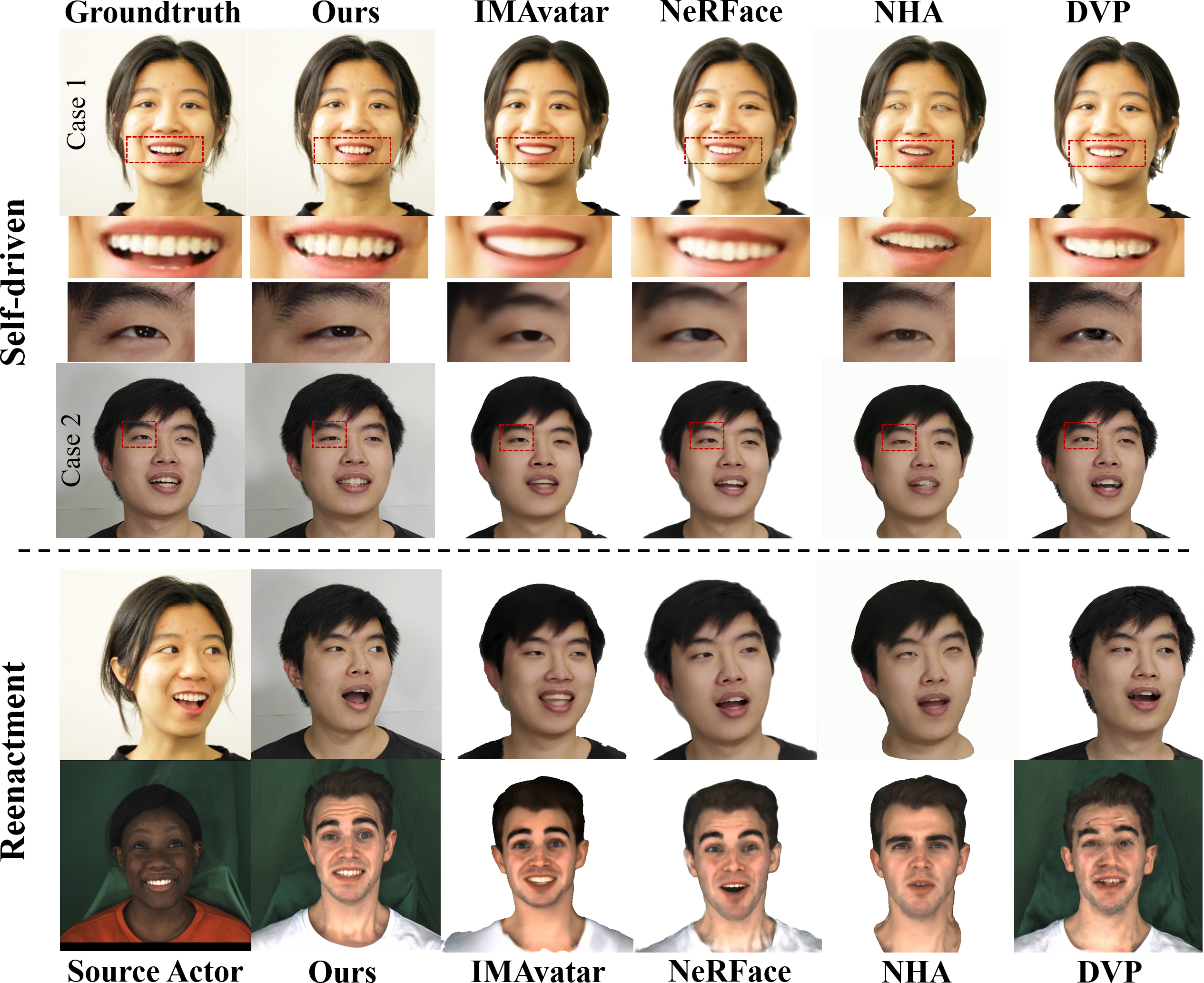}
	\end{center}
\caption{The comparisons with video-based facial reenactment methods on self-driven re-animation and faical reenactment. Our method can generate more realistic details (e.g. teeth, light points in eyes).} 
\label{fig:comparison}
\end{figure}

We compare our method with state-of-the-art video-based facial reenactment methods, including Deep Video Portrait (DVP)~\cite{kim2018deep}, NeRFace~\cite{nerface}, IMAvatar~\cite{zheng2022avatar}, and Neural Head Avatar (NHA)~\cite{grassal2021neuralhead}, all of which are trained on monocular videos. We have partitioned 80\% of the video into a training set and the remaining 20\% into a testing set for evaluation. Additionally, we have removed the background and resized the images to a resolution of $512\times512$. The values presented in Tab.~\ref{tab:tab2} are calculated based on the self-driving images and their corresponding ground-truth images in the testing set. We also perform self-driven re-animation and reenactment for these methods, as shown in Fig.~\ref{fig:comparison}. While the 3D methods show stable head geometry, they fail to produce high-fidelity texture details such as hair, teeth, and pupils in the output renderings. By incorporating StyleGAN-based networks and using our data augmentation, our method achieves higher image quality than the existing methods and can preserve more details such as light points in the eyes. As shown in Tab.~\ref{tab:tab2}, our method achieves significantly better quantitative results, particularly in the FID metric, indicating that our method generates higher-quality images.

\subsection{Ablation Study}
\label{sec:ablation}
\begin{figure}
	\begin{center}
		\includegraphics[width=\linewidth]{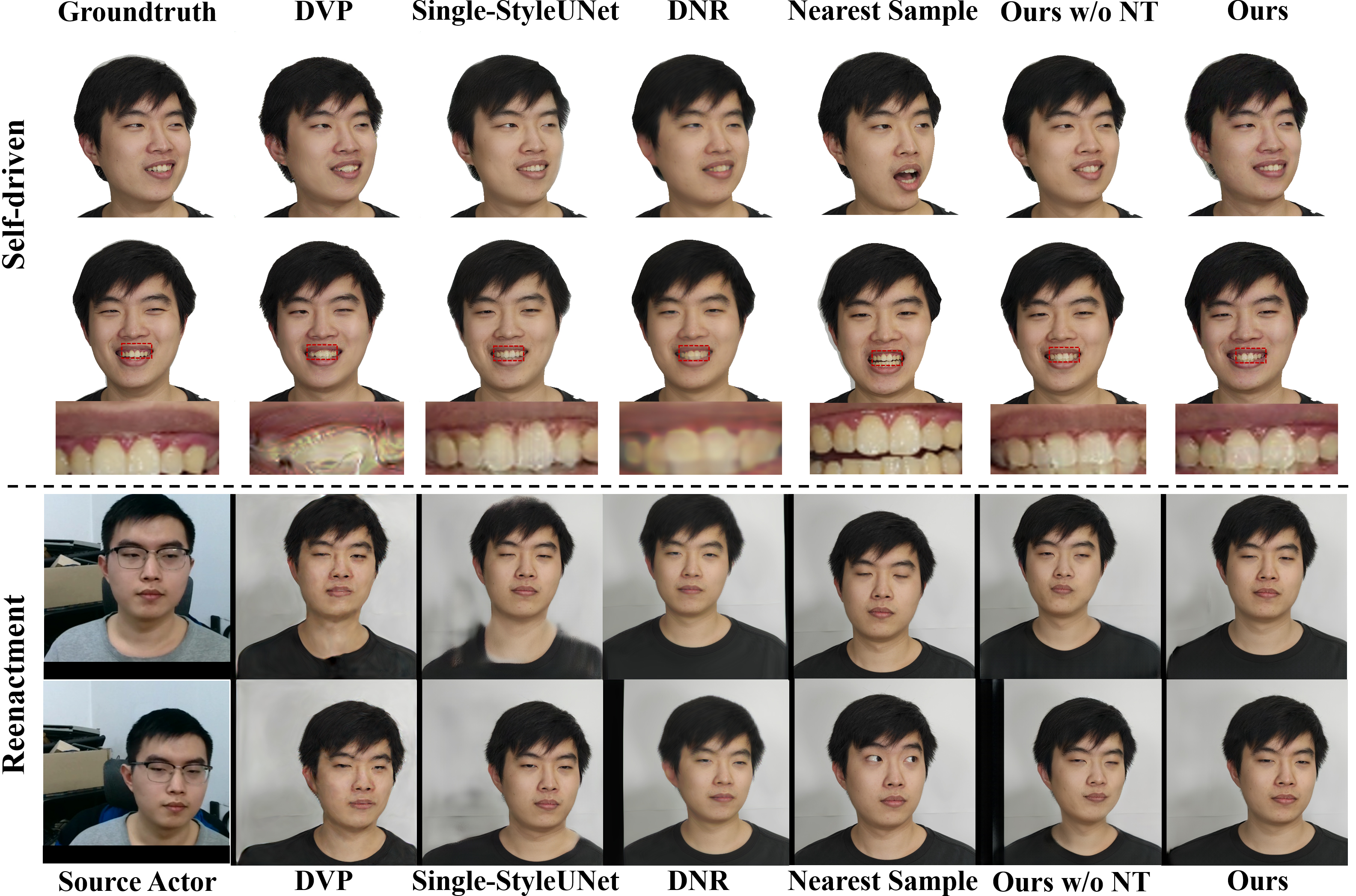}
	\end{center}
\caption{Ablation study of DVP, DNR, our ``single-StyleUNet'', ``nearest sample'', ``ours w/o NT'' and our full method. Our full method is superior in image quality and translation generalization.}
\label{fig:ablation}
\end{figure}

In our comparisons, we have demonstrated the powerful image generation capabilities of our network, which benefit from the StyleGAN-based StyleUNet structure. However, it should be noted that a simple StyleUNet may not be able to achieve translation generalization, and may require a long time to converge. In the ablation study, we use the term ``DVP (Ours)'' to refer to a simple UNet input with 3DMM texture rendering, similar to the DVP structure, and ``Single-StyleUNet'' to refer to a StyleUNet input with 3DMM texture rendering. ``DNR (Ours)'' represents a comparison with Deferred Neural Rendering~\cite{Thies2019DeferredNR}, which only uses the Neural Textures of the facial region as the input of a UNet. Additionally, we include ``Nearest Sample'' in our training set and ``Ours w/o NT'', which refers to our method without the Neural Textures.
As shown in Fig.~\ref{fig:ablation} and Tab.~\ref{tab:ablation}, without our data augmentation and video decomposition, even though ``Single-StyleUNet'' can still generate high-fidelity images, it is unable to prevent image tearing during significant head movements. Both ``DVP (Ours)'' and ``DNR (Ours)'' exhibit a noticeable decrease in image quality, highlighting the importance of StyleUNet in generating high-fidelity images. ``Ours w/o NT'' can still produce high-quality images, suggesting that the Neural Textures primarily contribute to speeding up the training convergence. Note that the cases 1 and 2 presented in Tab.~\ref{tab:ablation} have been marked in the self-driven portion of Fig.~\ref{fig:ablation}.

\begin{table}
\small
\begin{center}
\caption{Quantitative comparisons of our ablation study.}
    \begin{tabular}{lccc}
    \toprule
Error Metric   & SSIM $\uparrow$  & PSNR $\uparrow$  & FID $\downarrow$ \\ \midrule
DVP (Ours)   & 0.85 & 24.2 & 25.9   \\
Single-StyleUNet  & 0.87 & 26.2 & 15.05   \\
DNR (Ours)   & $\mathbf{0.89}$ & 26.1 & 27.2   \\
Nearest Sample  & 0.81 & 22.1 & 18.3   \\
Ours w/o NT  & 0.88 & $\mathbf{26.6}$ & 15.10   \\
Ours  & 0.87 & 26.2 & $\mathbf{13.2}$  \\ 
\bottomrule
    \end{tabular}
%\newline
\label{tab:ablation}
\end{center}
\end{table}

To validate the generalization ability of our method, we present the ``Nearest Sample'' in Fig.~\ref{fig:ablation} and calculate the average nearest translation, rotation, and expression parameters for both self-driven and reenactment cases, as shown in Tab.~\ref{tab:near}. It should be noted that we first normalize all parameters using the standard deviation calculated from the training set. The images in Fig.~\ref{fig:ablation} are selected with equal weight given to translation, rotation, and expression parameters, but the values in Tab.~\ref{tab:near} are calculated separately for each parameter. The results demonstrate that our method can achieve larger extrapolation on translation while performing similarly to other state-of-the-art methods on expression and rotation. Additionally, our method can only handle rotations up to approximately 30 degrees due to the challenges that large rotations present for face tracking.

\begin{table}
\small
\begin{center}
\caption{The average nearest parameters are presented for both the testing set and the reenactment case. Note that all parameters are normalized by the standard deviation calculated in the training set.}
    \begin{tabular}{lccc}
    \toprule
Parameter   & Translation $\uparrow$  & Rotation $\uparrow$  & Expression $\downarrow$ \\ \midrule
Testing Set   & 0.052 & 0.058 & 0.316   \\
Reenactment   & 2.002 & 0.054 & 0.294   \\
\bottomrule
    \end{tabular}
%\newline
\label{tab:near}
\end{center}
\end{table}

\begin{figure}
	\begin{center}
		\includegraphics[width=\linewidth]{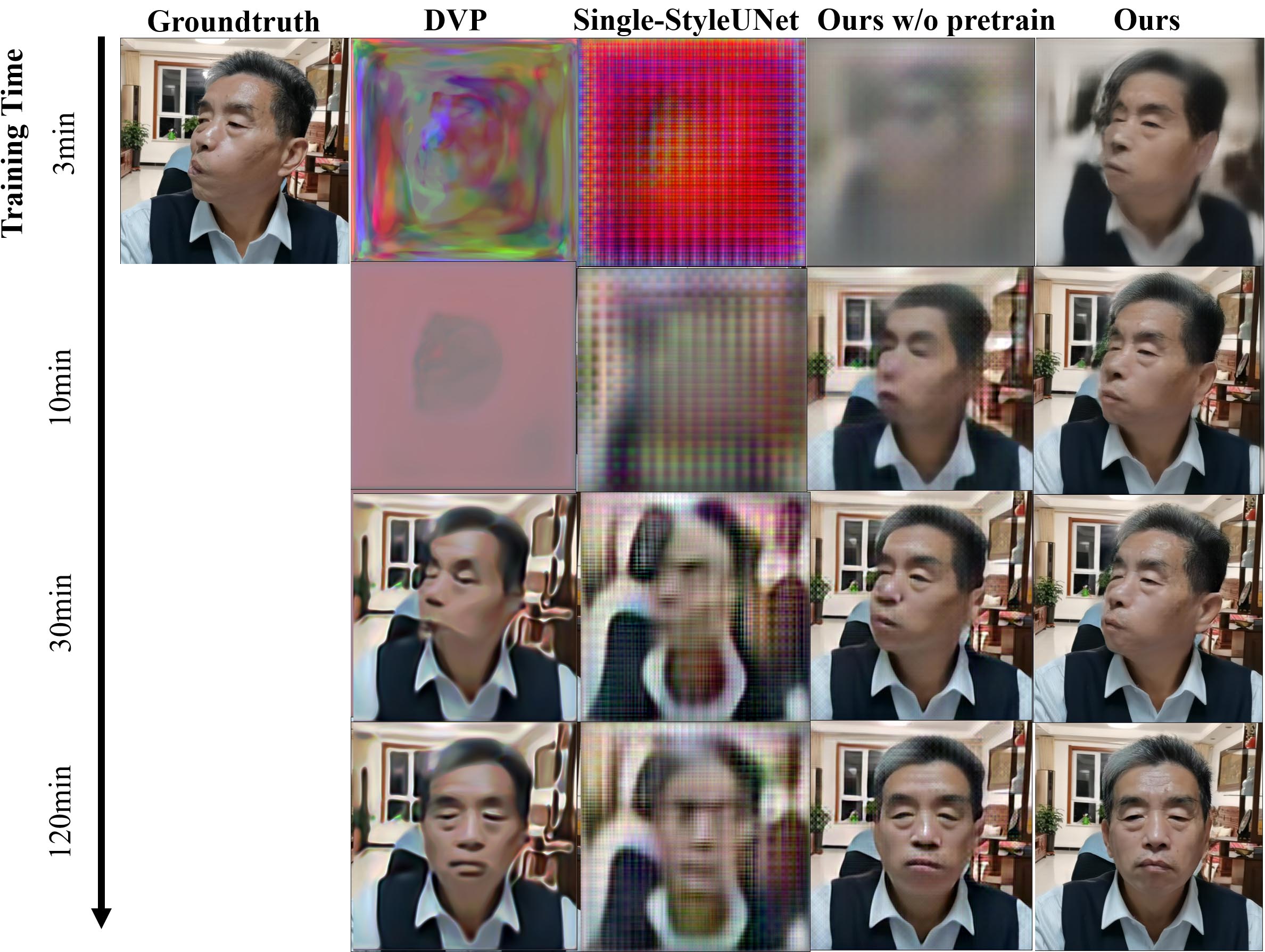}
	\end{center}
\caption{Generated images of different network structures during the training stage. Our method can achieve significantly faster training speed.}
\label{fig:train}
\end{figure}

As illustrated in Fig.\ref{fig:train}, due to the video decomposition and Neural Textures, our method achieves significantly faster training speed even without pre-training. The fourth column of ``Ours w/o pretrain'' can learn the basic information of the background and facial region faster than the third column. Only the remaining regions such as shoulders, hair, and small parts such as eyeglasses and teeth require further training. Comparing "Ours" with "Ours w/o pretrain" demonstrates the effectiveness of pre-training. Note that the video in Fig.\ref{fig:train} was not used in pre-training.

\subsection{Training and Rendering Efficiency}
\label{sec:eff}
As shown in Tab.~\ref{tab:tab3}, our method is significantly faster in training and rendering, while also generating images at a higher resolution. This can be attributed to our use of video decomposition, pre-training and a simple network structure that can be easily accelerated by TensorRT. Additionally, we use the vertex color of 3DMM points to represent their corresponding UV coordinates, allowing for real-time rendering using OpenGL. It should be noted that after just two hours of training, our model has already achieved a visually pleasing result, with only the teeth region requiring further training (approximately another 4 hours) to reach convergence.

\begin{table}
\footnotesize
\begin{center}
\caption{Training time, rendering time and image resolution of each method. Compared to the existing methods, our method is significantly faster in training and is able to render images at a higher resolution in real-time.}
    \begin{tabular}{lccc}
    \toprule
Method     & Training time (hour)   & Rendering time per frame (s)  & Resolution  \\ \midrule
IMAvatar  & ~48  & 100 & 512 \\
NeRFace   & ~36 & 4  & 512  \\
NHA       & ~13 & 0.06  & 512  \\
Ours      & ~2 & 0.028  & 1024 \\ \bottomrule
    \end{tabular}
%\newline
\label{tab:tab3}
\end{center}
\end{table}

\begin{figure}
	\begin{center}
		\includegraphics[width=0.9\linewidth]{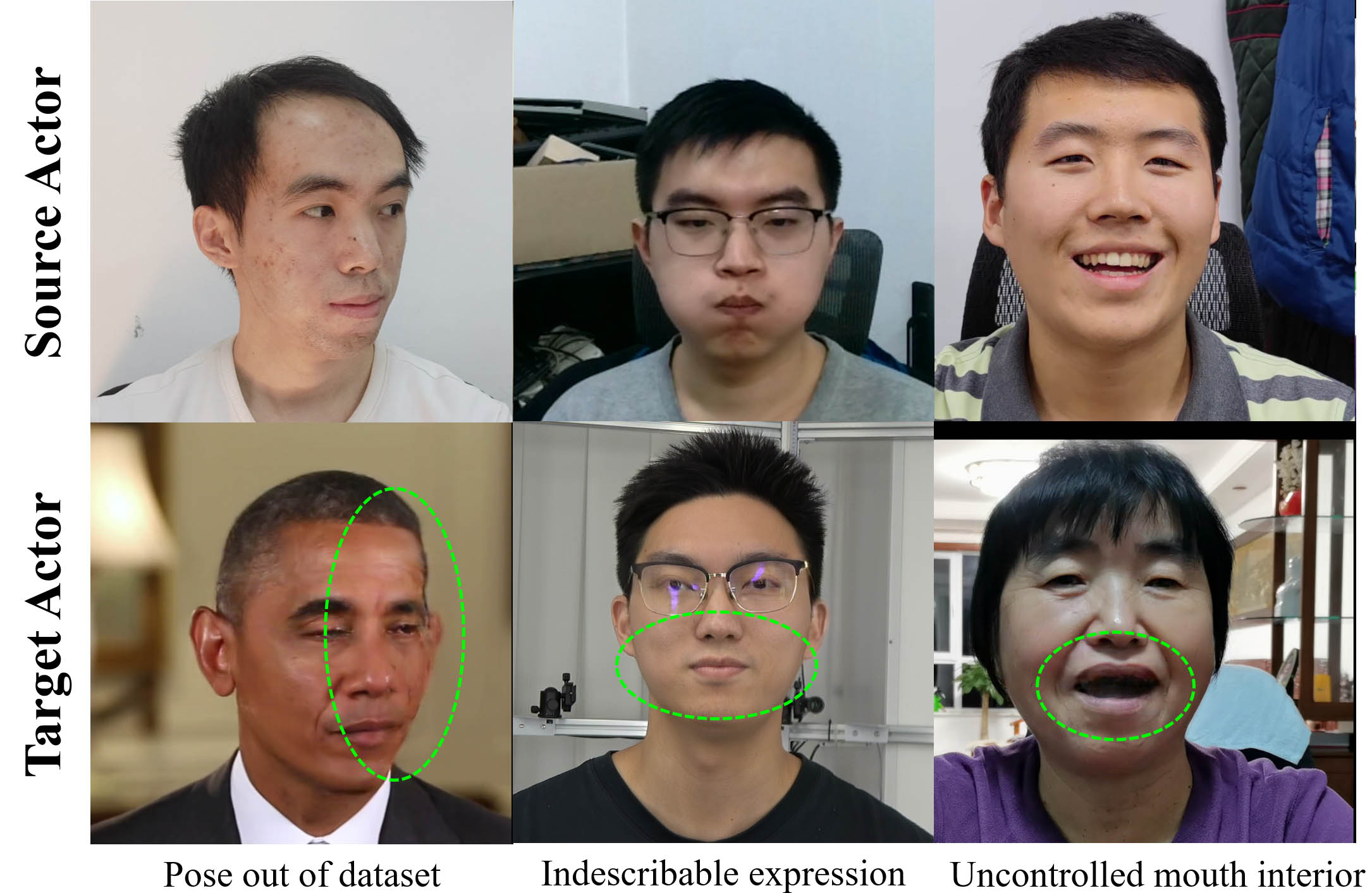}
	\end{center}
\caption{Failure cases arise due to poses outside of the training dataset, expressions that cannot be modeled by the parametric model, and uncontrollable mouth interior.}
\label{fig:limitation}
\end{figure}

\section{Discussion and Conclusion}
\label{sec:conclusion}

\paragraph{Limitations.} The proposed StyleAvatar outperforms state-of-the-art facial reenactment methods, but still has some limitations. First, our image-to-image translation network is limited by the quality and variation of the training dataset. As a result, we cannot generate rotations and expressions that differ significantly from the training dataset, as shown in the first column of Fig.~\ref{fig:limitation}. Second, the input renderings are generated by a 3DMM tracking algorithm. However, the tracked 3DMM is not capable of accurately describing detailed expressions, leading to inaccurate expression control, as shown in the second column of Fig.~\ref{fig:limitation}. Additionally, the mouth interior is not constrained, resulting in a lack of realism during the reenactment stage, with the inside of the mouth sometimes appearing blurred.

\paragraph{Potential Social Impact.} Our method enables a digital portrait copy that can be reenacted by another portrait video. Therefore, given a portrait video of a specific person, it can be used to generate fake portrait videos, which needs to be addressed carefully before deploying the technology.

\paragraph{Conclusion.} In this paper, we have presented StyleAvatar, a real-time photo-realistic portrait avatar generated from a single video. We have proposed a novel StyleGAN-based framework that can generate a full portrait video, including the shoulders and background, with high image quality. The unique video decomposition and the sliding window data augmentation enable us to achieve faster convergence and more natural movements. Additionally, our proposed live system allows the learned facial avatar to be re-animated by other subjects in real-time. Our extensive results and comprehensive experiments demonstrate that our method outperforms state-of-the-art methods for single-video-based facial avatar reconstruction and reenactment. We believe that our framework will inspire future research on facial reenactment, and our real-time live system has promising potential applications for related tasks.

\begin{acks}
This paper is supported by National Key R\&D Program of China (2022YFF0902200), the NSFC project No.62125107, No.61827805 and No.62171255, and Guoqiang Institute of Tsinghua University (No.2021GQG0001).
\end{acks}

\bibliographystyle{ACM-Reference-Format}
\bibliography{ref}


\begin{thebibliography}{77}

%%% the Web address.

\ifx \showCODEN    \undefined \def \showCODEN     #1{\unskip}     \fi
\ifx \showDOI      \undefined \def \showDOI       #1{#1}\fi
\ifx \showISBNx    \undefined \def \showISBNx     #1{\unskip}     \fi
\ifx \showISBNxiii \undefined \def \showISBNxiii  #1{\unskip}     \fi
\ifx \showISSN     \undefined \def \showISSN      #1{\unskip}     \fi
\ifx \showLCCN     \undefined \def \showLCCN      #1{\unskip}     \fi
\ifx \shownote     \undefined \def \shownote      #1{#1}          \fi
\ifx \showarticletitle \undefined \def \showarticletitle #1{#1}   \fi
\ifx \showURL      \undefined \def \showURL       {\relax}        \fi
% invisible to TeX
\providecommand\bibfield[2]{#2}
\providecommand\bibinfo[2]{#2}
\providecommand\natexlab[1]{#1}
\providecommand\showeprint[2][]{arXiv:#2}

\bibitem[Abdal et~al\mbox{.}(2021)]%
        {abdal2020styleflow}
\bibfield{author}{\bibinfo{person}{Rameen Abdal}, \bibinfo{person}{Peihao Zhu},
  \bibinfo{person}{Niloy~J Mitra}, {and} \bibinfo{person}{Peter Wonka}.}
  \bibinfo{year}{2021}\natexlab{}.
\newblock \showarticletitle{{StyleFlow}: Attribute-conditioned exploration of
  {StyleGAN-generated} images using conditional continuous normalizing flows}.
\newblock \bibinfo{journal}{\emph{ACM Transactions on Graphics (TOG)}}
  \bibinfo{volume}{40}, \bibinfo{number}{3} (\bibinfo{year}{2021}),
  \bibinfo{pages}{1--21}.
\newblock


\bibitem[Alaluf et~al\mbox{.}(2021)]%
        {alaluf2021restyle}
\bibfield{author}{\bibinfo{person}{Yuval Alaluf}, \bibinfo{person}{Or
  Patashnik}, {and} \bibinfo{person}{Daniel Cohen-Or}.}
  \bibinfo{year}{2021}\natexlab{}.
\newblock \showarticletitle{Restyle: A residual-based {StyleGAN} encoder via
  iterative refinement}. In \bibinfo{booktitle}{\emph{IEEE/CVF International
  Conference on Computer Vision (ICCV)}}. \bibinfo{pages}{6711--6720}.
\newblock


\bibitem[Averbuch-Elor et~al\mbox{.}(2017)]%
        {averbuch-elor2017bringing}
\bibfield{author}{\bibinfo{person}{Hadar Averbuch-Elor},
  \bibinfo{person}{Daniel Cohen-Or}, \bibinfo{person}{Johannes Kopf}, {and}
  \bibinfo{person}{Michael~F Cohen}.} \bibinfo{year}{2017}\natexlab{}.
\newblock \showarticletitle{Bringing portraits to life}.
\newblock \bibinfo{journal}{\emph{ACM transactions on graphics (TOG)}}
  \bibinfo{volume}{36}, \bibinfo{number}{6} (\bibinfo{year}{2017}),
  \bibinfo{pages}{1--13}.
\newblock


\bibitem[Bahmani et~al\mbox{.}(2022)]%
        {bahmani20223d}
\bibfield{author}{\bibinfo{person}{Sherwin Bahmani},
  \bibinfo{person}{Jeong~Joon Park}, \bibinfo{person}{Despoina Paschalidou},
  \bibinfo{person}{Hao Tang}, \bibinfo{person}{Gordon Wetzstein},
  \bibinfo{person}{Leonidas Guibas}, \bibinfo{person}{Luc Van~Gool}, {and}
  \bibinfo{person}{Radu Timofte}.} \bibinfo{year}{2022}\natexlab{}.
\newblock \showarticletitle{3d-aware video generation}.
\newblock \bibinfo{journal}{\emph{arXiv preprint arXiv:2206.14797}}
  (\bibinfo{year}{2022}).
\newblock


\bibitem[Blanz and Vetter(1999)]%
        {CGIT1999Blanz}
\bibfield{author}{\bibinfo{person}{Volker Blanz} {and} \bibinfo{person}{Thomas
  Vetter}.} \bibinfo{year}{1999}\natexlab{}.
\newblock \showarticletitle{A Morphable Model for the Synthesis of {3D} Faces}.
  In \bibinfo{booktitle}{\emph{ACM SIGGRAPH}}. \bibinfo{publisher}{ACM},
  \bibinfo{pages}{187--194}.
\newblock


\bibitem[Cao et~al\mbox{.}(2022)]%
        {cao2022authentic}
\bibfield{author}{\bibinfo{person}{Chen Cao}, \bibinfo{person}{Tomas Simon},
  \bibinfo{person}{Jin~Kyu Kim}, \bibinfo{person}{Gabe Schwartz},
  \bibinfo{person}{Michael Zollhoefer}, \bibinfo{person}{Shun-Suke Saito},
  \bibinfo{person}{Stephen Lombardi}, \bibinfo{person}{Shih-En Wei},
  \bibinfo{person}{Danielle Belko}, \bibinfo{person}{Shoou-I Yu},
  {et~al\mbox{.}}} \bibinfo{year}{2022}\natexlab{}.
\newblock \showarticletitle{Authentic volumetric avatars from a phone scan}.
\newblock \bibinfo{journal}{\emph{ACM Transactions on Graphics (TOG)}}
  \bibinfo{volume}{41}, \bibinfo{number}{4} (\bibinfo{year}{2022}),
  \bibinfo{pages}{1--19}.
\newblock


\bibitem[Chan et~al\mbox{.}(2022)]%
        {Chan2022eg3d}
\bibfield{author}{\bibinfo{person}{Eric~R Chan}, \bibinfo{person}{Connor~Z
  Lin}, \bibinfo{person}{Matthew~A Chan}, \bibinfo{person}{Koki Nagano},
  \bibinfo{person}{Boxiao Pan}, \bibinfo{person}{Shalini De~Mello},
  \bibinfo{person}{Orazio Gallo}, \bibinfo{person}{Leonidas~J Guibas},
  \bibinfo{person}{Jonathan Tremblay}, \bibinfo{person}{Sameh Khamis},
  {et~al\mbox{.}}} \bibinfo{year}{2022}\natexlab{}.
\newblock \showarticletitle{Efficient geometry-aware {3D} generative
  adversarial networks}. In \bibinfo{booktitle}{\emph{IEEE/CVF Conference on
  Computer Vision and Pattern Recognition (CVPR)}}.
  \bibinfo{pages}{16123--16133}.
\newblock


\bibitem[Chen et~al\mbox{.}(2022)]%
        {sofgan}
\bibfield{author}{\bibinfo{person}{Anpei Chen}, \bibinfo{person}{Ruiyang Liu},
  \bibinfo{person}{Ling Xie}, \bibinfo{person}{Zhang Chen},
  \bibinfo{person}{Hao Su}, {and} \bibinfo{person}{Jingyi Yu}.}
  \bibinfo{year}{2022}\natexlab{}.
\newblock \showarticletitle{{SofGAN}: A portrait image generator with dynamic
  styling}.
\newblock \bibinfo{journal}{\emph{ACM Transactions on Graphics (TOG)}}
  \bibinfo{volume}{41}, \bibinfo{number}{1} (\bibinfo{year}{2022}),
  \bibinfo{pages}{1--26}.
\newblock


\bibitem[Deng et~al\mbox{.}(2020)]%
        {deng2020disentangled}
\bibfield{author}{\bibinfo{person}{Yu Deng}, \bibinfo{person}{Jiaolong Yang},
  \bibinfo{person}{Dong Chen}, \bibinfo{person}{Fang Wen}, {and}
  \bibinfo{person}{Xin Tong}.} \bibinfo{year}{2020}\natexlab{}.
\newblock \showarticletitle{Disentangled and controllable face image generation
  via {3D} imitative-contrastive learning}. In
  \bibinfo{booktitle}{\emph{IEEE/CVF Conference on Computer Vision and Pattern
  Recognition (CVPR)}}. \bibinfo{pages}{5154--5163}.
\newblock


\bibitem[Doukas et~al\mbox{.}(2021a)]%
        {doukas2021head2head}
\bibfield{author}{\bibinfo{person}{Michail~Christos Doukas},
  \bibinfo{person}{Mohammad~Rami Koujan}, \bibinfo{person}{Viktoriia
  Sharmanska}, \bibinfo{person}{Anastasios Roussos}, {and}
  \bibinfo{person}{Stefanos Zafeiriou}.} \bibinfo{year}{2021}\natexlab{a}.
\newblock \showarticletitle{{Head2Head++}: Deep facial attributes
  re-targeting}.
\newblock \bibinfo{journal}{\emph{IEEE Transactions on Biometrics, Behavior,
  and Identity Science}} \bibinfo{volume}{3}, \bibinfo{number}{1}
  (\bibinfo{year}{2021}), \bibinfo{pages}{31--43}.
\newblock


\bibitem[Doukas et~al\mbox{.}(2021b)]%
        {doukas2021headgan}
\bibfield{author}{\bibinfo{person}{Michail~Christos Doukas},
  \bibinfo{person}{Stefanos Zafeiriou}, {and} \bibinfo{person}{Viktoriia
  Sharmanska}.} \bibinfo{year}{2021}\natexlab{b}.
\newblock \showarticletitle{{HeadGAN}: One-shot neural head synthesis and
  editing}. In \bibinfo{booktitle}{\emph{IEEE/CVF International Conference on
  Computer Vision (ICCV)}}. \bibinfo{pages}{14398--14407}.
\newblock


\bibitem[Drobyshev et~al\mbox{.}(2022)]%
        {Drobyshev22MP}
\bibfield{author}{\bibinfo{person}{Nikita Drobyshev}, \bibinfo{person}{Jenya
  Chelishev}, \bibinfo{person}{Taras Khakhulin}, \bibinfo{person}{Aleksei
  Ivakhnenko}, \bibinfo{person}{Victor Lempitsky}, {and} \bibinfo{person}{Egor
  Zakharov}.} \bibinfo{year}{2022}\natexlab{}.
\newblock \showarticletitle{{MegaPortraits}: One-shot megapixel neural head
  avatars}.
\newblock \bibinfo{journal}{\emph{arXiv preprint arXiv:2207.07621}}
  (\bibinfo{year}{2022}).
\newblock


\bibitem[Fang et~al\mbox{.}(2022)]%
        {TiNeuVox}
\bibfield{author}{\bibinfo{person}{Jiemin Fang}, \bibinfo{person}{Taoran Yi},
  \bibinfo{person}{Xinggang Wang}, \bibinfo{person}{Lingxi Xie},
  \bibinfo{person}{Xiaopeng Zhang}, \bibinfo{person}{Wenyu Liu},
  \bibinfo{person}{Matthias Nie{\ss}ner}, {and} \bibinfo{person}{Qi Tian}.}
  \bibinfo{year}{2022}\natexlab{}.
\newblock \showarticletitle{Fast dynamic radiance fields with time-aware neural
  voxels}. In \bibinfo{booktitle}{\emph{SIGGRAPH Asia 2022 Conference Papers}}.
  \bibinfo{pages}{1--9}.
\newblock


\bibitem[Gafni et~al\mbox{.}(2021)]%
        {nerface}
\bibfield{author}{\bibinfo{person}{Guy Gafni}, \bibinfo{person}{Justus Thies},
  \bibinfo{person}{Michael Zollhofer}, {and} \bibinfo{person}{Matthias
  Nie{\ss}ner}.} \bibinfo{year}{2021}\natexlab{}.
\newblock \showarticletitle{Dynamic neural radiance fields for monocular {4D}
  facial avatar reconstruction}. In \bibinfo{booktitle}{\emph{IEEE/CVF
  Conference on Computer Vision and Pattern Recognition (CVPR)}}.
  \bibinfo{pages}{8649--8658}.
\newblock


\bibitem[Gal et~al\mbox{.}(2021)]%
        {gal2021swagan}
\bibfield{author}{\bibinfo{person}{Rinon Gal}, \bibinfo{person}{Dana~Cohen
  Hochberg}, \bibinfo{person}{Amit Bermano}, {and} \bibinfo{person}{Daniel
  Cohen-Or}.} \bibinfo{year}{2021}\natexlab{}.
\newblock \showarticletitle{{SWAGAN}: A style-based wavelet-driven generative
  model}.
\newblock \bibinfo{journal}{\emph{ACM Transactions on Graphics (TOG)}}
  \bibinfo{volume}{40}, \bibinfo{number}{4} (\bibinfo{year}{2021}),
  \bibinfo{pages}{1--11}.
\newblock


\bibitem[Gao et~al\mbox{.}(2022)]%
        {Gao2022nerfblendshape}
\bibfield{author}{\bibinfo{person}{Xuan Gao}, \bibinfo{person}{Chenglai Zhong},
  \bibinfo{person}{Jun Xiang}, \bibinfo{person}{Yang Hong},
  \bibinfo{person}{Yudong Guo}, {and} \bibinfo{person}{Juyong Zhang}.}
  \bibinfo{year}{2022}\natexlab{}.
\newblock \showarticletitle{Reconstructing personalized semantic facial {NeRF}
  models from monocular video}.
\newblock \bibinfo{journal}{\emph{ACM Transactions on Graphics (TOG)}}
  \bibinfo{volume}{41}, \bibinfo{number}{6} (\bibinfo{year}{2022}),
  \bibinfo{pages}{1--12}.
\newblock


\bibitem[Garbin et~al\mbox{.}(2022)]%
        {garbin2022voltemorph}
\bibfield{author}{\bibinfo{person}{Stephan~J Garbin}, \bibinfo{person}{Marek
  Kowalski}, \bibinfo{person}{Virginia Estellers}, \bibinfo{person}{Stanislaw
  Szymanowicz}, \bibinfo{person}{Shideh Rezaeifar}, \bibinfo{person}{Jingjing
  Shen}, \bibinfo{person}{Matthew Johnson}, {and} \bibinfo{person}{Julien
  Valentin}.} \bibinfo{year}{2022}\natexlab{}.
\newblock \showarticletitle{{VolTeMorph}: Realtime, Controllable and
  Generalisable Animation of Volumetric Representations}.
\newblock \bibinfo{journal}{\emph{arXiv preprint arXiv:2208.00949}}
  (\bibinfo{year}{2022}).
\newblock


\bibitem[Garrido et~al\mbox{.}(2014)]%
        {garrido2014automatic}
\bibfield{author}{\bibinfo{person}{Pablo Garrido}, \bibinfo{person}{Levi
  Valgaerts}, \bibinfo{person}{Ole Rehmsen}, \bibinfo{person}{Thorsten
  Thormahlen}, \bibinfo{person}{Patrick Perez}, {and}
  \bibinfo{person}{Christian Theobalt}.} \bibinfo{year}{2014}\natexlab{}.
\newblock \showarticletitle{Automatic face reenactment}. In
  \bibinfo{booktitle}{\emph{IEEE Conference on Computer Vision and Pattern
  Recognition (CVPR)}}. \bibinfo{pages}{4217--4224}.
\newblock


\bibitem[Geng et~al\mbox{.}(2018)]%
        {geng2019warp}
\bibfield{author}{\bibinfo{person}{Jiahao Geng}, \bibinfo{person}{Tianjia
  Shao}, \bibinfo{person}{Youyi Zheng}, \bibinfo{person}{Yanlin Weng}, {and}
  \bibinfo{person}{Kun Zhou}.} \bibinfo{year}{2018}\natexlab{}.
\newblock \showarticletitle{Warp-guided {GANs} for single-photo facial
  animation}.
\newblock \bibinfo{journal}{\emph{ACM Transactions on Graphics (TOG)}}
  \bibinfo{volume}{37}, \bibinfo{number}{6} (\bibinfo{year}{2018}),
  \bibinfo{pages}{1--12}.
\newblock


\bibitem[Ghosh et~al\mbox{.}(2020)]%
        {ghosh2020gif}
\bibfield{author}{\bibinfo{person}{Partha Ghosh}, \bibinfo{person}{Pravir~Singh
  Gupta}, \bibinfo{person}{Roy Uziel}, \bibinfo{person}{Anurag Ranjan},
  \bibinfo{person}{Michael~J Black}, {and} \bibinfo{person}{Timo Bolkart}.}
  \bibinfo{year}{2020}\natexlab{}.
\newblock \showarticletitle{{GIF}: Generative interpretable faces}. In
  \bibinfo{booktitle}{\emph{International Conference on 3D Vision (3DV)}}.
  \bibinfo{publisher}{IEEE}, \bibinfo{pages}{868--878}.
\newblock


\bibitem[Grassal et~al\mbox{.}(2022a)]%
        {grassal2021neuralhead}
\bibfield{author}{\bibinfo{person}{Philip-William Grassal},
  \bibinfo{person}{Malte Prinzler}, \bibinfo{person}{Titus Leistner},
  \bibinfo{person}{Carsten Rother}, \bibinfo{person}{Matthias Nie{\ss}ner},
  {and} \bibinfo{person}{Justus Thies}.} \bibinfo{year}{2022}\natexlab{a}.
\newblock \showarticletitle{Neural head avatars from monocular {RGB} videos}.
  In \bibinfo{booktitle}{\emph{IEEE/CVF Conference on Computer Vision and
  Pattern Recognition (CVPR)}}. \bibinfo{pages}{18653--18664}.
\newblock


\bibitem[Grassal et~al\mbox{.}(2022b)]%
        {grassal2022neural}
\bibfield{author}{\bibinfo{person}{Philip-William Grassal},
  \bibinfo{person}{Malte Prinzler}, \bibinfo{person}{Titus Leistner},
  \bibinfo{person}{Carsten Rother}, \bibinfo{person}{Matthias Nie{\ss}ner},
  {and} \bibinfo{person}{Justus Thies}.} \bibinfo{year}{2022}\natexlab{b}.
\newblock \showarticletitle{Neural head avatars from monocular {RGB} videos}.
  In \bibinfo{booktitle}{\emph{IEEE/CVF Conference on Computer Vision and
  Pattern Recognition (CVPR)}}. \bibinfo{pages}{18653--18664}.
\newblock


\bibitem[H{\"a}rk{\"o}nen et~al\mbox{.}(2020)]%
        {harkonen2020ganspace}
\bibfield{author}{\bibinfo{person}{Erik H{\"a}rk{\"o}nen},
  \bibinfo{person}{Aaron Hertzmann}, \bibinfo{person}{Jaakko Lehtinen}, {and}
  \bibinfo{person}{Sylvain Paris}.} \bibinfo{year}{2020}\natexlab{}.
\newblock \showarticletitle{{GANSpace}: Discovering interpretable {GAN}
  controls}.
\newblock \bibinfo{journal}{\emph{Advances in Neural Information Processing
  Systems}}  \bibinfo{volume}{33} (\bibinfo{year}{2020}),
  \bibinfo{pages}{9841--9850}.
\newblock


\bibitem[Hong et~al\mbox{.}(2022)]%
        {hong2022depth}
\bibfield{author}{\bibinfo{person}{Fa-Ting Hong}, \bibinfo{person}{Longhao
  Zhang}, \bibinfo{person}{Li Shen}, {and} \bibinfo{person}{Dan Xu}.}
  \bibinfo{year}{2022}\natexlab{}.
\newblock \showarticletitle{Depth-aware generative adversarial network for
  talking head video generation}. In \bibinfo{booktitle}{\emph{IEEE/CVF
  Conference on Computer Vision and Pattern Recognition (CVPR)}}.
  \bibinfo{pages}{3397--3406}.
\newblock


\bibitem[Jang et~al\mbox{.}(2021)]%
        {jang2021stylecarigan}
\bibfield{author}{\bibinfo{person}{Wonjong Jang}, \bibinfo{person}{Gwangjin
  Ju}, \bibinfo{person}{Yucheol Jung}, \bibinfo{person}{Jiaolong Yang},
  \bibinfo{person}{Xin Tong}, {and} \bibinfo{person}{Seungyong Lee}.}
  \bibinfo{year}{2021}\natexlab{}.
\newblock \showarticletitle{{StyleCariGAN}: caricature generation via
  {StyleGAN} feature map modulation}.
\newblock \bibinfo{journal}{\emph{ACM Transactions on Graphics (TOG)}}
  \bibinfo{volume}{40}, \bibinfo{number}{4} (\bibinfo{year}{2021}),
  \bibinfo{pages}{1--16}.
\newblock


\bibitem[Karras et~al\mbox{.}(2021)]%
        {styleganv3}
\bibfield{author}{\bibinfo{person}{Tero Karras}, \bibinfo{person}{Miika
  Aittala}, \bibinfo{person}{Samuli Laine}, \bibinfo{person}{Erik
  H{\"a}rk{\"o}nen}, \bibinfo{person}{Janne Hellsten}, \bibinfo{person}{Jaakko
  Lehtinen}, {and} \bibinfo{person}{Timo Aila}.}
  \bibinfo{year}{2021}\natexlab{}.
\newblock \showarticletitle{Alias-free generative adversarial networks}.
\newblock \bibinfo{journal}{\emph{Advances in Neural Information Processing
  Systems}}  \bibinfo{volume}{34} (\bibinfo{year}{2021}),
  \bibinfo{pages}{852--863}.
\newblock


\bibitem[Karras et~al\mbox{.}(2019)]%
        {karras2020a}
\bibfield{author}{\bibinfo{person}{Tero Karras}, \bibinfo{person}{Samuli
  Laine}, {and} \bibinfo{person}{Timo Aila}.} \bibinfo{year}{2019}\natexlab{}.
\newblock \showarticletitle{A style-based generator architecture for generative
  adversarial networks}. In \bibinfo{booktitle}{\emph{IEEE/CVF Conference on
  Computer Vision and Pattern Recognition (CVPR)}}.
  \bibinfo{pages}{4401--4410}.
\newblock


\bibitem[Karras et~al\mbox{.}(2020)]%
        {karras2020analyzing}
\bibfield{author}{\bibinfo{person}{Tero Karras}, \bibinfo{person}{Samuli
  Laine}, \bibinfo{person}{Miika Aittala}, \bibinfo{person}{Janne Hellsten},
  \bibinfo{person}{Jaakko Lehtinen}, {and} \bibinfo{person}{Timo Aila}.}
  \bibinfo{year}{2020}\natexlab{}.
\newblock \showarticletitle{Analyzing and improving the image quality of
  {StyleGAN}}. In \bibinfo{booktitle}{\emph{IEEE/CVF Conference on Computer
  Vision and Pattern Recognition (CVPR)}}. \bibinfo{pages}{8110--8119}.
\newblock


\bibitem[Khakhulin et~al\mbox{.}(2022)]%
        {Khakhulin2022ROME}
\bibfield{author}{\bibinfo{person}{Taras Khakhulin}, \bibinfo{person}{Vanessa
  Sklyarova}, \bibinfo{person}{Victor Lempitsky}, {and} \bibinfo{person}{Egor
  Zakharov}.} \bibinfo{year}{2022}\natexlab{}.
\newblock \showarticletitle{Realistic one-shot mesh-based head avatars}. In
  \bibinfo{booktitle}{\emph{European Conference of Computer vision (ECCV)}}.
  \bibinfo{publisher}{Springer}, \bibinfo{pages}{345--362}.
\newblock


\bibitem[Kim et~al\mbox{.}(2018)]%
        {kim2018deep}
\bibfield{author}{\bibinfo{person}{Hyeongwoo Kim}, \bibinfo{person}{Pablo
  Garrido}, \bibinfo{person}{Ayush Tewari}, \bibinfo{person}{Weipeng Xu},
  \bibinfo{person}{Justus Thies}, \bibinfo{person}{Matthias Niessner},
  \bibinfo{person}{Patrick P{\'e}rez}, \bibinfo{person}{Christian Richardt},
  \bibinfo{person}{Michael Zollh{\"o}fer}, {and} \bibinfo{person}{Christian
  Theobalt}.} \bibinfo{year}{2018}\natexlab{}.
\newblock \showarticletitle{Deep video portraits}.
\newblock \bibinfo{journal}{\emph{ACM Transactions on Graphics (TOG)}}
  \bibinfo{volume}{37}, \bibinfo{number}{4} (\bibinfo{year}{2018}),
  \bibinfo{pages}{1--14}.
\newblock


\bibitem[Koujan et~al\mbox{.}(2020)]%
        {koujan2020head2head}
\bibfield{author}{\bibinfo{person}{Mohammad~Rami Koujan},
  \bibinfo{person}{Michail~Christos Doukas}, \bibinfo{person}{Anastasios
  Roussos}, {and} \bibinfo{person}{Stefanos Zafeiriou}.}
  \bibinfo{year}{2020}\natexlab{}.
\newblock \showarticletitle{Head2head: Video-based neural head synthesis}. In
  \bibinfo{booktitle}{\emph{IEEE International Conference on Automatic Face and
  Gesture Recognition}}. \bibinfo{publisher}{IEEE}, \bibinfo{pages}{16--23}.
\newblock


\bibitem[Kowalski et~al\mbox{.}(2020)]%
        {kowalski2020config}
\bibfield{author}{\bibinfo{person}{Marek Kowalski}, \bibinfo{person}{Stephan~J
  Garbin}, \bibinfo{person}{Virginia Estellers}, \bibinfo{person}{Tadas
  Baltru{\v{s}}aitis}, \bibinfo{person}{Matthew Johnson}, {and}
  \bibinfo{person}{Jamie Shotton}.} \bibinfo{year}{2020}\natexlab{}.
\newblock \showarticletitle{Config: Controllable neural face image generation}.
  In \bibinfo{booktitle}{\emph{European Conference on Computer Vision (ECCV)}}.
  \bibinfo{publisher}{Springer}, \bibinfo{pages}{299--315}.
\newblock


\bibitem[Li et~al\mbox{.}(2012)]%
        {li2012data}
\bibfield{author}{\bibinfo{person}{Kai Li}, \bibinfo{person}{Feng Xu},
  \bibinfo{person}{Jue Wang}, \bibinfo{person}{Qionghai Dai}, {and}
  \bibinfo{person}{Yebin Liu}.} \bibinfo{year}{2012}\natexlab{}.
\newblock \showarticletitle{A data-driven approach for facial expression
  synthesis in video}. In \bibinfo{booktitle}{\emph{2012 IEEE Conference on
  Computer Vision and Pattern Recognition}}. IEEE, \bibinfo{pages}{57--64}.
\newblock


\bibitem[Lin et~al\mbox{.}(2021)]%
        {rvm}
\bibfield{author}{\bibinfo{person}{Shanchuan Lin}, \bibinfo{person}{Linjie
  Yang}, \bibinfo{person}{Imran Saleemi}, {and} \bibinfo{person}{Soumyadip
  Sengupta}.} \bibinfo{year}{2021}\natexlab{}.
\newblock \showarticletitle{Robust High-Resolution Video Matting with Temporal
  Guidance}.
\newblock \bibinfo{journal}{\emph{arXiv preprint arXiv:2108.11515}}
  (\bibinfo{year}{2021}).
\newblock


\bibitem[Liu et~al\mbox{.}(2001)]%
        {liu2001expressive}
\bibfield{author}{\bibinfo{person}{Zicheng Liu}, \bibinfo{person}{Ying Shan},
  {and} \bibinfo{person}{Zhengyou Zhang}.} \bibinfo{year}{2001}\natexlab{}.
\newblock \showarticletitle{Expressive expression mapping with ratio images}.
  In \bibinfo{booktitle}{\emph{Proceedings of the 28th Annual Conference on
  Computer Graphics and Interactive Techniques}}. \bibinfo{pages}{271--276}.
\newblock


\bibitem[Lombardi et~al\mbox{.}(2018)]%
        {lombardi2018deep}
\bibfield{author}{\bibinfo{person}{Stephen Lombardi}, \bibinfo{person}{Jason
  Saragih}, \bibinfo{person}{Tomas Simon}, {and} \bibinfo{person}{Yaser
  Sheikh}.} \bibinfo{year}{2018}\natexlab{}.
\newblock \showarticletitle{Deep appearance models for face rendering}.
\newblock \bibinfo{journal}{\emph{ACM Transactions on Graphics (TOG)}}
  \bibinfo{volume}{37}, \bibinfo{number}{4} (\bibinfo{year}{2018}),
  \bibinfo{pages}{1--13}.
\newblock


\bibitem[Lombardi et~al\mbox{.}(2019)]%
        {lombardi2019neural}
\bibfield{author}{\bibinfo{person}{Stephen Lombardi}, \bibinfo{person}{Tomas
  Simon}, \bibinfo{person}{Jason Saragih}, \bibinfo{person}{Gabriel Schwartz},
  \bibinfo{person}{Andreas Lehrmann}, {and} \bibinfo{person}{Yaser Sheikh}.}
  \bibinfo{year}{2019}\natexlab{}.
\newblock \showarticletitle{Neural volumes: Learning dynamic renderable volumes
  from images}.
\newblock \bibinfo{journal}{\emph{arXiv preprint arXiv:1906.07751}}
  (\bibinfo{year}{2019}).
\newblock


\bibitem[Ma et~al\mbox{.}(2021)]%
        {ma2021pixel}
\bibfield{author}{\bibinfo{person}{Shugao Ma}, \bibinfo{person}{Tomas Simon},
  \bibinfo{person}{Jason Saragih}, \bibinfo{person}{Dawei Wang},
  \bibinfo{person}{Yuecheng Li}, \bibinfo{person}{Fernando De~La~Torre}, {and}
  \bibinfo{person}{Yaser Sheikh}.} \bibinfo{year}{2021}\natexlab{}.
\newblock \showarticletitle{Pixel codec avatars}. In
  \bibinfo{booktitle}{\emph{IEEE/CVF Conference on Computer Vision and Pattern
  Recognition (CVPR)}}. \bibinfo{pages}{64--73}.
\newblock


\bibitem[Mallya et~al\mbox{.}(2022)]%
        {mallya2022implicit}
\bibfield{author}{\bibinfo{person}{Arun Mallya}, \bibinfo{person}{Ting-Chun
  Wang}, {and} \bibinfo{person}{Ming-Yu Liu}.} \bibinfo{year}{2022}\natexlab{}.
\newblock \showarticletitle{Implicit Warping for Animation with Image Sets}.
\newblock \bibinfo{journal}{\emph{arXiv preprint arXiv:2210.01794}}
  (\bibinfo{year}{2022}).
\newblock


\bibitem[M{\"u}ller et~al\mbox{.}(2022)]%
        {mueller2022instantngp}
\bibfield{author}{\bibinfo{person}{Thomas M{\"u}ller}, \bibinfo{person}{Alex
  Evans}, \bibinfo{person}{Christoph Schied}, {and} \bibinfo{person}{Alexander
  Keller}.} \bibinfo{year}{2022}\natexlab{}.
\newblock \showarticletitle{Instant neural graphics primitives with a
  multiresolution hash encoding}.
\newblock \bibinfo{journal}{\emph{ACM Transactions on Graphics (TOG)}}
  \bibinfo{volume}{41}, \bibinfo{number}{4} (\bibinfo{year}{2022}),
  \bibinfo{pages}{1--15}.
\newblock


\bibitem[Nagano et~al\mbox{.}(2018)]%
        {nagano2019pagan}
\bibfield{author}{\bibinfo{person}{Koki Nagano}, \bibinfo{person}{Jaewoo Seo},
  \bibinfo{person}{Jun Xing}, \bibinfo{person}{Lingyu Wei},
  \bibinfo{person}{Zimo Li}, \bibinfo{person}{Shunsuke Saito},
  \bibinfo{person}{Aviral Agarwal}, \bibinfo{person}{Jens Fursund},
  \bibinfo{person}{Hao Li}, \bibinfo{person}{Richard Roberts}, {et~al\mbox{.}}}
  \bibinfo{year}{2018}\natexlab{}.
\newblock \showarticletitle{{paGAN}: real-time avatars using dynamic textures.}
\newblock \bibinfo{journal}{\emph{ACM Transactions on Graphics (TOG)}}
  \bibinfo{volume}{37}, \bibinfo{number}{6} (\bibinfo{year}{2018}),
  \bibinfo{pages}{258--1}.
\newblock


\bibitem[Olszewski et~al\mbox{.}(2017)]%
        {olszewski2017realistic}
\bibfield{author}{\bibinfo{person}{Kyle Olszewski}, \bibinfo{person}{Zimo Li},
  \bibinfo{person}{Chao Yang}, \bibinfo{person}{Yi Zhou},
  \bibinfo{person}{Ronald Yu}, \bibinfo{person}{Zeng Huang},
  \bibinfo{person}{Sitao Xiang}, \bibinfo{person}{Shunsuke Saito},
  \bibinfo{person}{Pushmeet Kohli}, {and} \bibinfo{person}{Hao Li}.}
  \bibinfo{year}{2017}\natexlab{}.
\newblock \showarticletitle{Realistic dynamic facial textures from a single
  image using {GANs}}. In \bibinfo{booktitle}{\emph{IEEE International
  Conference on Computer Vision (ICCV)}}. \bibinfo{pages}{5429--5438}.
\newblock


\bibitem[Perov et~al\mbox{.}(2020)]%
        {deepfacelab}
\bibfield{author}{\bibinfo{person}{Ivan Perov}, \bibinfo{person}{Daiheng Gao},
  \bibinfo{person}{Nikolay Chervoniy}, \bibinfo{person}{Kunlin Liu},
  \bibinfo{person}{Sugasa Marangonda}, \bibinfo{person}{Chris Um{\'e}},
  \bibinfo{person}{Mr Dpfks}, \bibinfo{person}{Carl~Shift Facenheim},
  \bibinfo{person}{Luis RP}, \bibinfo{person}{Jian Jiang}, {et~al\mbox{.}}}
  \bibinfo{year}{2020}\natexlab{}.
\newblock \showarticletitle{DeepFaceLab: Integrated, flexible and extensible
  face-swapping framework}.
\newblock \bibinfo{journal}{\emph{arXiv preprint arXiv:2005.05535}}
  (\bibinfo{year}{2020}).
\newblock


\bibitem[Raj et~al\mbox{.}(2021)]%
        {raj2021pva}
\bibfield{author}{\bibinfo{person}{Amit Raj}, \bibinfo{person}{Michael
  Zollhoefer}, \bibinfo{person}{Tomas Simon}, \bibinfo{person}{Jason Saragih},
  \bibinfo{person}{Shunsuke Saito}, \bibinfo{person}{James Hays}, {and}
  \bibinfo{person}{Stephen Lombardi}.} \bibinfo{year}{2021}\natexlab{}.
\newblock \showarticletitle{{PVA}: Pixel-aligned volumetric avatars}.
\newblock \bibinfo{journal}{\emph{arXiv preprint arXiv:2101.02697}}
  (\bibinfo{year}{2021}).
\newblock


\bibitem[Ren et~al\mbox{.}(2021)]%
        {ren2021pirenderer}
\bibfield{author}{\bibinfo{person}{Yurui Ren}, \bibinfo{person}{Ge Li},
  \bibinfo{person}{Yuanqi Chen}, \bibinfo{person}{Thomas~H Li}, {and}
  \bibinfo{person}{Shan Liu}.} \bibinfo{year}{2021}\natexlab{}.
\newblock \showarticletitle{Pirenderer: Controllable portrait image generation
  via semantic neural rendering}. In \bibinfo{booktitle}{\emph{IEEE/CVF
  International Conference on Computer Vision (ICCV)}}.
  \bibinfo{pages}{13759--13768}.
\newblock


\bibitem[Richardson et~al\mbox{.}(2021)]%
        {richardson2021encoding}
\bibfield{author}{\bibinfo{person}{Elad Richardson}, \bibinfo{person}{Yuval
  Alaluf}, \bibinfo{person}{Or Patashnik}, \bibinfo{person}{Yotam Nitzan},
  \bibinfo{person}{Yaniv Azar}, \bibinfo{person}{Stav Shapiro}, {and}
  \bibinfo{person}{Daniel Cohen-Or}.} \bibinfo{year}{2021}\natexlab{}.
\newblock \showarticletitle{Encoding in style: a {StyleGAN} encoder for
  image-to-image translation}. In \bibinfo{booktitle}{\emph{IEEE/CVF Conference
  on Computer Vision and Pattern Recognition (CVPR)}}.
  \bibinfo{pages}{2287--2296}.
\newblock


\bibitem[Ronneberger et~al\mbox{.}(2015)]%
        {ronneberger2015u-net:}
\bibfield{author}{\bibinfo{person}{Olaf Ronneberger}, \bibinfo{person}{Philipp
  Fischer}, {and} \bibinfo{person}{Thomas Brox}.}
  \bibinfo{year}{2015}\natexlab{}.
\newblock \showarticletitle{{U-Net}: Convolutional networks for biomedical
  image segmentation}. In \bibinfo{booktitle}{\emph{Medical Image Computing and
  Computer-Assisted Intervention (MICCAI)}}. \bibinfo{publisher}{Springer},
  \bibinfo{pages}{234--241}.
\newblock


\bibitem[Shen et~al\mbox{.}(2020a)]%
        {shen2020interpreting}
\bibfield{author}{\bibinfo{person}{Yujun Shen}, \bibinfo{person}{Jinjin Gu},
  \bibinfo{person}{Xiaoou Tang}, {and} \bibinfo{person}{Bolei Zhou}.}
  \bibinfo{year}{2020}\natexlab{a}.
\newblock \showarticletitle{Interpreting the latent space of {GANs} for
  semantic face editing}. In \bibinfo{booktitle}{\emph{IEEE/CVF Conference on
  Computer Vision and Pattern Recognition (CVPR)}}.
  \bibinfo{pages}{9243--9252}.
\newblock


\bibitem[Shen et~al\mbox{.}(2020b)]%
        {shen2020interfacegan}
\bibfield{author}{\bibinfo{person}{Yujun Shen}, \bibinfo{person}{Ceyuan Yang},
  \bibinfo{person}{Xiaoou Tang}, {and} \bibinfo{person}{Bolei Zhou}.}
  \bibinfo{year}{2020}\natexlab{b}.
\newblock \showarticletitle{{InterFaceGAN}: Interpreting the disentangled face
  representation learned by {GANs}}.
\newblock \bibinfo{journal}{\emph{IEEE Transactions on Pattern Analysis and
  Machine Intelligence (TPAMI)}} \bibinfo{volume}{44}, \bibinfo{number}{4}
  (\bibinfo{year}{2020}), \bibinfo{pages}{2004--2018}.
\newblock


\bibitem[Shi et~al\mbox{.}(2022)]%
        {shi2022semanticstylegan}
\bibfield{author}{\bibinfo{person}{Yichun Shi}, \bibinfo{person}{Xiao Yang},
  \bibinfo{person}{Yangyue Wan}, {and} \bibinfo{person}{Xiaohui Shen}.}
  \bibinfo{year}{2022}\natexlab{}.
\newblock \showarticletitle{{SemanticStyleGAN}: Learning compositional
  generative priors for controllable image synthesis and editing}. In
  \bibinfo{booktitle}{\emph{IEEE/CVF Conference on Computer Vision and Pattern
  Recognition (CVPR)}}. \bibinfo{pages}{11254--11264}.
\newblock


\bibitem[Shoshan et~al\mbox{.}(2021)]%
        {shoshan2021gan}
\bibfield{author}{\bibinfo{person}{Alon Shoshan}, \bibinfo{person}{Nadav
  Bhonker}, \bibinfo{person}{Igor Kviatkovsky}, {and} \bibinfo{person}{Gerard
  Medioni}.} \bibinfo{year}{2021}\natexlab{}.
\newblock \showarticletitle{{GAN-Control}: Explicitly controllable {GANs}}. In
  \bibinfo{booktitle}{\emph{IEEE/CVF International Conference on Computer
  Vision (ICCV)}}. \bibinfo{pages}{14083--14093}.
\newblock


\bibitem[Siarohin et~al\mbox{.}(2019)]%
        {siarohin2019first}
\bibfield{author}{\bibinfo{person}{Aliaksandr Siarohin},
  \bibinfo{person}{St{\'e}phane Lathuili{\`e}re}, \bibinfo{person}{Sergey
  Tulyakov}, \bibinfo{person}{Elisa Ricci}, {and} \bibinfo{person}{Nicu Sebe}.}
  \bibinfo{year}{2019}\natexlab{}.
\newblock \showarticletitle{First order motion model for image animation}.
\newblock \bibinfo{journal}{\emph{Advances in Neural Information Processing
  Systems}}  \bibinfo{volume}{32} (\bibinfo{year}{2019}).
\newblock


\bibitem[Sun et~al\mbox{.}(2022)]%
        {sun2022ide}
\bibfield{author}{\bibinfo{person}{Jingxiang Sun}, \bibinfo{person}{Xuan Wang},
  \bibinfo{person}{Yichun Shi}, \bibinfo{person}{Lizhen Wang},
  \bibinfo{person}{Jue Wang}, {and} \bibinfo{person}{Yebin Liu}.}
  \bibinfo{year}{2022}\natexlab{}.
\newblock \showarticletitle{{IDE-3D}: Interactive disentangled editing for
  High-Resolution {3D}-Aware portrait synthesis}.
\newblock \bibinfo{journal}{\emph{ACM Transactions on Graphics (TOG)}}
  \bibinfo{volume}{41}, \bibinfo{number}{6} (\bibinfo{year}{2022}),
  \bibinfo{pages}{1--10}.
\newblock


\bibitem[Sun et~al\mbox{.}(2023)]%
        {sun2022next3d}
\bibfield{author}{\bibinfo{person}{Jingxiang Sun}, \bibinfo{person}{Xuan Wang},
  \bibinfo{person}{Lizhen Wang}, \bibinfo{person}{Xiaoyu Li},
  \bibinfo{person}{Yong Zhang}, \bibinfo{person}{Hongwen Zhang}, {and}
  \bibinfo{person}{Yebin Liu}.} \bibinfo{year}{2023}\natexlab{}.
\newblock \showarticletitle{Next3D: Generative Neural Texture Rasterization for
  3D-Aware Head Avatars}. In \bibinfo{booktitle}{\emph{IEEE/CVF Conference on
  Computer Vision and Pattern Recognition (CVPR)}}.
\newblock


\bibitem[Suwajanakorn et~al\mbox{.}(2017)]%
        {suwajanakorn2017synthesizing}
\bibfield{author}{\bibinfo{person}{Supasorn Suwajanakorn},
  \bibinfo{person}{Steven~M Seitz}, {and} \bibinfo{person}{Ira
  Kemelmacher-Shlizerman}.} \bibinfo{year}{2017}\natexlab{}.
\newblock \showarticletitle{Synthesizing obama: learning lip sync from audio}.
\newblock \bibinfo{journal}{\emph{ACM Transactions on Graphics (TOG)}}
  \bibinfo{volume}{36}, \bibinfo{number}{4} (\bibinfo{year}{2017}),
  \bibinfo{pages}{1--13}.
\newblock


\bibitem[Tewari et~al\mbox{.}(2020a)]%
        {tewari2020pie}
\bibfield{author}{\bibinfo{person}{Ayush Tewari}, \bibinfo{person}{Mohamed
  Elgharib}, \bibinfo{person}{Florian Bernard}, \bibinfo{person}{Hans-Peter
  Seidel}, \bibinfo{person}{Patrick P{\'e}rez}, \bibinfo{person}{Michael
  Zollh{\"o}fer}, {and} \bibinfo{person}{Christian Theobalt}.}
  \bibinfo{year}{2020}\natexlab{a}.
\newblock \showarticletitle{{PIE}: Portrait image embedding for semantic
  control}.
\newblock \bibinfo{journal}{\emph{ACM Transactions on Graphics (TOG)}}
  \bibinfo{volume}{39}, \bibinfo{number}{6} (\bibinfo{year}{2020}),
  \bibinfo{pages}{1--14}.
\newblock


\bibitem[Tewari et~al\mbox{.}(2020b)]%
        {tewari2020stylerig}
\bibfield{author}{\bibinfo{person}{Ayush Tewari}, \bibinfo{person}{Mohamed
  Elgharib}, \bibinfo{person}{Gaurav Bharaj}, \bibinfo{person}{Florian
  Bernard}, \bibinfo{person}{Hans-Peter Seidel}, \bibinfo{person}{Patrick
  P{\'e}rez}, \bibinfo{person}{Michael Zollhofer}, {and}
  \bibinfo{person}{Christian Theobalt}.} \bibinfo{year}{2020}\natexlab{b}.
\newblock \showarticletitle{{StyleRig}: Rigging {StyleGAN} for {3D} control
  over portrait images}. In \bibinfo{booktitle}{\emph{IEEE/CVF Conference on
  Computer Vision and Pattern Recognition (CVPR)}}.
  \bibinfo{pages}{6142--6151}.
\newblock


\bibitem[Thies et~al\mbox{.}(2019)]%
        {Thies2019DeferredNR}
\bibfield{author}{\bibinfo{person}{Justus Thies}, \bibinfo{person}{Michael
  Zollh{\"o}fer}, {and} \bibinfo{person}{Matthias Nie{\ss}ner}.}
  \bibinfo{year}{2019}\natexlab{}.
\newblock \showarticletitle{Deferred neural rendering: Image synthesis using
  neural textures}.
\newblock \bibinfo{journal}{\emph{Acm Transactions on Graphics (TOG)}}
  \bibinfo{volume}{38}, \bibinfo{number}{4} (\bibinfo{year}{2019}),
  \bibinfo{pages}{1--12}.
\newblock


\bibitem[Thies et~al\mbox{.}(2015)]%
        {thies2015real}
\bibfield{author}{\bibinfo{person}{Justus Thies}, \bibinfo{person}{Michael
  Zollh{\"o}fer}, \bibinfo{person}{Matthias Nie{\ss}ner}, \bibinfo{person}{Levi
  Valgaerts}, \bibinfo{person}{Marc Stamminger}, {and}
  \bibinfo{person}{Christian Theobalt}.} \bibinfo{year}{2015}\natexlab{}.
\newblock \showarticletitle{Real-time expression transfer for facial
  reenactment.}
\newblock \bibinfo{journal}{\emph{ACM Transactions on Graphics (TOG)}}
  \bibinfo{volume}{34}, \bibinfo{number}{6} (\bibinfo{year}{2015}),
  \bibinfo{pages}{183--1}.
\newblock


\bibitem[Thies et~al\mbox{.}(2016)]%
        {thies2016face2face}
\bibfield{author}{\bibinfo{person}{Justus Thies}, \bibinfo{person}{Michael
  Zollhofer}, \bibinfo{person}{Marc Stamminger}, \bibinfo{person}{Christian
  Theobalt}, {and} \bibinfo{person}{Matthias Nie{\ss}ner}.}
  \bibinfo{year}{2016}\natexlab{}.
\newblock \showarticletitle{{Face2Face}: Real-time face capture and reenactment
  of {RGB} videos}. In \bibinfo{booktitle}{\emph{IEEE Conference on Computer
  Vision and Pattern Recognition (CVPR)}}. \bibinfo{pages}{2387--2395}.
\newblock


\bibitem[Tov et~al\mbox{.}(2021)]%
        {tov2021designing}
\bibfield{author}{\bibinfo{person}{Omer Tov}, \bibinfo{person}{Yuval Alaluf},
  \bibinfo{person}{Yotam Nitzan}, \bibinfo{person}{Or Patashnik}, {and}
  \bibinfo{person}{Daniel Cohen-Or}.} \bibinfo{year}{2021}\natexlab{}.
\newblock \showarticletitle{Designing an encoder for {StyleGAN} image
  manipulation}.
\newblock \bibinfo{journal}{\emph{ACM Transactions on Graphics (TOG)}}
  \bibinfo{volume}{40}, \bibinfo{number}{4} (\bibinfo{year}{2021}),
  \bibinfo{pages}{1--14}.
\newblock


\bibitem[Vlasic et~al\mbox{.}(2005)]%
        {vlasic2005face}
\bibfield{author}{\bibinfo{person}{Daniel Vlasic}, \bibinfo{person}{Matthew
  Brand}, \bibinfo{person}{Hanspeter Pfister}, {and} \bibinfo{person}{Jovan
  Popovi{\'c}}.} \bibinfo{year}{2005}\natexlab{}.
\newblock \showarticletitle{Face transfer with multilinear models}.
\newblock \bibinfo{journal}{\emph{ACM Transactions on Graphics (TOG)}}
  \bibinfo{volume}{24}, \bibinfo{number}{3} (\bibinfo{year}{2005}),
  \bibinfo{pages}{426--433}.
\newblock


\bibitem[Wang et~al\mbox{.}(2022a)]%
        {Wang2022MoRFMR}
\bibfield{author}{\bibinfo{person}{Daoye Wang}, \bibinfo{person}{Prashanth
  Chandran}, \bibinfo{person}{Gaspard Zoss}, \bibinfo{person}{Derek Bradley},
  {and} \bibinfo{person}{Paulo Gotardo}.} \bibinfo{year}{2022}\natexlab{a}.
\newblock \showarticletitle{{MoRF}: Morphable radiance fields for multiview
  neural head modeling}. In \bibinfo{booktitle}{\emph{ACM SIGGRAPH 2022
  Conference Proceedings}}. \bibinfo{pages}{1--9}.
\newblock


\bibitem[Wang et~al\mbox{.}(2020)]%
        {kaisiyuan2020mead}
\bibfield{author}{\bibinfo{person}{Kaisiyuan Wang}, \bibinfo{person}{Qianyi
  Wu}, \bibinfo{person}{Linsen Song}, \bibinfo{person}{Zhuoqian Yang},
  \bibinfo{person}{Wayne Wu}, \bibinfo{person}{Chen Qian}, \bibinfo{person}{Ran
  He}, \bibinfo{person}{Yu Qiao}, {and} \bibinfo{person}{Chen~Change Loy}.}
  \bibinfo{year}{2020}\natexlab{}.
\newblock \showarticletitle{{MEAD}: A large-scale audio-visual dataset for
  emotional talking-face generation}. In \bibinfo{booktitle}{\emph{European
  Conference on Computer Vision (ECCV)}}. \bibinfo{publisher}{Springer},
  \bibinfo{pages}{700--717}.
\newblock


\bibitem[Wang et~al\mbox{.}(2022b)]%
        {wang2022faceverse}
\bibfield{author}{\bibinfo{person}{Lizhen Wang}, \bibinfo{person}{Zhiyuan
  Chen}, \bibinfo{person}{Tao Yu}, \bibinfo{person}{Chenguang Ma},
  \bibinfo{person}{Liang Li}, {and} \bibinfo{person}{Yebin Liu}.}
  \bibinfo{year}{2022}\natexlab{b}.
\newblock \showarticletitle{{FaceVerse}: a fine-grained and detail-controllable
  {3D} face morphable model from a hybrid dataset}. In
  \bibinfo{booktitle}{\emph{IEEE/CVF Conference on Computer Vision and Pattern
  Recognition (CVPR)}}. \bibinfo{pages}{20333--20342}.
\newblock


\bibitem[Wang et~al\mbox{.}(2021b)]%
        {Wang_2021_CVPR}
\bibfield{author}{\bibinfo{person}{Xintao Wang}, \bibinfo{person}{Yu Li},
  \bibinfo{person}{Honglun Zhang}, {and} \bibinfo{person}{Ying Shan}.}
  \bibinfo{year}{2021}\natexlab{b}.
\newblock \showarticletitle{Towards Real-World blind face restoration with
  generative facial prior}. In \bibinfo{booktitle}{\emph{IEEE/CVF Conference on
  Computer Vision and Pattern Recognition (CVPR)}}.
  \bibinfo{pages}{9168--9178}.
\newblock


\bibitem[Wang et~al\mbox{.}(2021a)]%
        {wang2021learning}
\bibfield{author}{\bibinfo{person}{Ziyan Wang}, \bibinfo{person}{Timur
  Bagautdinov}, \bibinfo{person}{Stephen Lombardi}, \bibinfo{person}{Tomas
  Simon}, \bibinfo{person}{Jason Saragih}, \bibinfo{person}{Jessica Hodgins},
  {and} \bibinfo{person}{Michael Zollhofer}.} \bibinfo{year}{2021}\natexlab{a}.
\newblock \showarticletitle{Learning Compositional Radiance Fields of Dynamic
  Human Heads}. In \bibinfo{booktitle}{\emph{IEEE/CVF Conference on Computer
  Vision and Pattern Recognition (CVPR)}}. \bibinfo{pages}{5704--5713}.
\newblock


\bibitem[Wei et~al\mbox{.}(2019)]%
        {wei2019vr}
\bibfield{author}{\bibinfo{person}{Shih-En Wei}, \bibinfo{person}{Jason
  Saragih}, \bibinfo{person}{Tomas Simon}, \bibinfo{person}{Adam~W Harley},
  \bibinfo{person}{Stephen Lombardi}, \bibinfo{person}{Michal Perdoch},
  \bibinfo{person}{Alexander Hypes}, \bibinfo{person}{Dawei Wang},
  \bibinfo{person}{Hernan Badino}, {and} \bibinfo{person}{Yaser Sheikh}.}
  \bibinfo{year}{2019}\natexlab{}.
\newblock \showarticletitle{{VR} facial animation via multiview image
  translation}.
\newblock \bibinfo{journal}{\emph{ACM Transactions on Graphics (TOG)}}
  \bibinfo{volume}{38}, \bibinfo{number}{4} (\bibinfo{year}{2019}),
  \bibinfo{pages}{1--16}.
\newblock


\bibitem[Weise et~al\mbox{.}(2011)]%
        {weise2011realtime}
\bibfield{author}{\bibinfo{person}{Thibaut Weise}, \bibinfo{person}{Sofien
  Bouaziz}, \bibinfo{person}{Hao Li}, {and} \bibinfo{person}{Mark Pauly}.}
  \bibinfo{year}{2011}\natexlab{}.
\newblock \showarticletitle{Realtime performance-based facial animation}.
\newblock \bibinfo{journal}{\emph{ACM transactions on graphics (TOG)}}
  \bibinfo{volume}{30}, \bibinfo{number}{4} (\bibinfo{year}{2011}),
  \bibinfo{pages}{1--10}.
\newblock


\bibitem[Xiang et~al\mbox{.}(2022)]%
        {xiang2022gram}
\bibfield{author}{\bibinfo{person}{Jianfeng Xiang}, \bibinfo{person}{Jiaolong
  Yang}, \bibinfo{person}{Yu Deng}, {and} \bibinfo{person}{Xin Tong}.}
  \bibinfo{year}{2022}\natexlab{}.
\newblock \showarticletitle{{GRAM-HD}: {3D}-Consistent image generation at high
  resolution with generative radiance manifolds}.
\newblock \bibinfo{journal}{\emph{arXiv preprint arXiv:2206.07255}}
  (\bibinfo{year}{2022}).
\newblock


\bibitem[Xu et~al\mbox{.}(2023a)]%
        {xu2023avatarmav}
\bibfield{author}{\bibinfo{person}{Yuelang Xu}, \bibinfo{person}{Lizhen Wang},
  \bibinfo{person}{Xiaochen Zhao}, \bibinfo{person}{Hongwen Zhang}, {and}
  \bibinfo{person}{Yebin Liu}.} \bibinfo{year}{2023}\natexlab{a}.
\newblock \showarticletitle{AvatarMAV: Fast 3D Head Avatar Reconstruction Using
  Motion-Aware Neural Voxels}. In \bibinfo{booktitle}{\emph{ACM SIGGRAPH 2023
  Conference Proceedings}}.
\newblock


\bibitem[Xu et~al\mbox{.}(2023b)]%
        {xu2023latentavatar}
\bibfield{author}{\bibinfo{person}{Yuelang Xu}, \bibinfo{person}{Hongwen
  Zhang}, \bibinfo{person}{Lizhen Wang}, \bibinfo{person}{Xiaochen Zhao},
  \bibinfo{person}{Han Huang}, \bibinfo{person}{Guojun Qi}, {and}
  \bibinfo{person}{Yebin Liu}.} \bibinfo{year}{2023}\natexlab{b}.
\newblock \showarticletitle{LatentAvatar: Learning Latent Expression Code for
  Expressive Neural Head Avatar}. In \bibinfo{booktitle}{\emph{ACM SIGGRAPH
  2023 Conference Proceedings}}.
\newblock


\bibitem[Yin et~al\mbox{.}(2022)]%
        {yin2022styleheat}
\bibfield{author}{\bibinfo{person}{Fei Yin}, \bibinfo{person}{Yong Zhang},
  \bibinfo{person}{Xiaodong Cun}, \bibinfo{person}{Mingdeng Cao},
  \bibinfo{person}{Yanbo Fan}, \bibinfo{person}{Xuan Wang},
  \bibinfo{person}{Qingyan Bai}, \bibinfo{person}{Baoyuan Wu},
  \bibinfo{person}{Jue Wang}, {and} \bibinfo{person}{Yujiu Yang}.}
  \bibinfo{year}{2022}\natexlab{}.
\newblock \showarticletitle{{StyleHEAT}: One-shot high-resolution editable
  talking face generation via pre-trained {StyleGAN}}. In
  \bibinfo{booktitle}{\emph{European Conference on Computer Vision (ECCV)}}.
  \bibinfo{publisher}{Springer}, \bibinfo{pages}{85--101}.
\newblock


\bibitem[Zheng et~al\mbox{.}(2022a)]%
        {zheng2022avatar}
\bibfield{author}{\bibinfo{person}{Yufeng Zheng},
  \bibinfo{person}{Victoria~Fern{\'a}ndez Abrevaya}, \bibinfo{person}{Marcel~C
  B{\"u}hler}, \bibinfo{person}{Xu Chen}, \bibinfo{person}{Michael~J Black},
  {and} \bibinfo{person}{Otmar Hilliges}.} \bibinfo{year}{2022}\natexlab{a}.
\newblock \showarticletitle{{IM} {Avatar}: Implicit morphable head avatars from
  videos}. In \bibinfo{booktitle}{\emph{IEEE/CVF Conference on Computer Vision
  and Pattern Recognition (CVPR)}}. \bibinfo{pages}{13545--13555}.
\newblock


\bibitem[Zheng et~al\mbox{.}(2022b)]%
        {zheng2022pointavatar}
\bibfield{author}{\bibinfo{person}{Yufeng Zheng}, \bibinfo{person}{Wang Yifan},
  \bibinfo{person}{Gordon Wetzstein}, \bibinfo{person}{Michael~J Black}, {and}
  \bibinfo{person}{Otmar Hilliges}.} \bibinfo{year}{2022}\natexlab{b}.
\newblock \showarticletitle{{PointAvatar}: Deformable Point-based Head Avatars
  from Videos}.
\newblock \bibinfo{journal}{\emph{arXiv preprint arXiv:2212.08377}}
  (\bibinfo{year}{2022}).
\newblock


\bibitem[Zheng et~al\mbox{.}(2023)]%
        {zheng2023avatarrex}
\bibfield{author}{\bibinfo{person}{Zerong Zheng}, \bibinfo{person}{Xiaochen
  Zhao}, \bibinfo{person}{Hongwen Zhang}, \bibinfo{person}{Boning Liu}, {and}
  \bibinfo{person}{Yebin Liu}.} \bibinfo{year}{2023}\natexlab{}.
\newblock \showarticletitle{AvatarReX: Real-time Expressive Full-body Avatars}.
\newblock \bibinfo{journal}{\emph{ACM Transactions on Graphics (TOG)}}
  \bibinfo{volume}{42}, \bibinfo{number}{4} (\bibinfo{year}{2023}),
  \bibinfo{pages}{1--19}.
\newblock
\urldef\tempurl%
\url{https://doi.org/10.1145/3592101}
\showDOI{\tempurl}


\bibitem[Zielonka et~al\mbox{.}(2022)]%
        {zielonka2022instant}
\bibfield{author}{\bibinfo{person}{Wojciech Zielonka}, \bibinfo{person}{Timo
  Bolkart}, {and} \bibinfo{person}{Justus Thies}.}
  \bibinfo{year}{2022}\natexlab{}.
\newblock \showarticletitle{Instant Volumetric Head Avatars}.
\newblock \bibinfo{journal}{\emph{arXiv preprint arXiv:2211.12499}}
  (\bibinfo{year}{2022}).
\newblock


\end{thebibliography}

%\newpage

%\appendix
%\input{tex/supp}

\end{document}